\documentclass[twoside]{IEEEtran}
\usepackage{graphicx,amssymb,amsmath,multirow,float,cite,subfigure,CJK,booktabs,mathtools} 
\usepackage[colorlinks,
            linkcolor=blue,
            anchorcolor=blue,
            citecolor=blue]{hyperref}

\ifCLASSINFOpdf
  
\else
 
\fi

\begin{document}
%
\title{A Scene-Text Synthesis Engine Achieved Through Learning from Decomposed Real-World Data}  
%
%

\author{Zhengmi~Tang,~\IEEEmembership{Member,~IEEE,}
        Tomo~Miyazaki,~\IEEEmembership{Member,~IEEE,}
        and~Shinichiro~Omachi,~\IEEEmembership{Senior Member,~IEEE}


\thanks{This work was supported in part by the Japan Society for the Promotion of Science (JSPS) KAKENHI under Grant
19K12033 and Grant 22H00540. 

The authors are with the Graduate School of Engineering, Tohoku University, Sendai, 980-8579, Japan. (E-mail: tzm@dc.tohoku.ac.jp, tomo@tohoku.ac.jp, machi@ecei.tohoku.ac.jp)}

}


\maketitle

\begin{abstract}
Scene-text image synthesis techniques that aim to naturally compose text instances on background scene images are very appealing for training deep neural networks due to their ability to provide accurate and comprehensive annotation information. Prior studies have explored generating synthetic text images on two-dimensional and three-dimensional surfaces using rules derived from real-world observations. Some of these studies have proposed generating scene-text images through learning; however, owing to the absence of a suitable training dataset, unsupervised frameworks have been explored to learn from existing real-world data, which might not yield reliable performance. To ease this dilemma and facilitate research on learning-based scene text synthesis, we introduce DecompST, a real-world dataset prepared from some public benchmarks, containing three types of annotations: quadrilateral-level BBoxes, stroke-level text masks, and text-erased images. Leveraging the DecompST dataset, we propose a Learning-Based Text Synthesis engine (LBTS) that includes a text location proposal network (TLPNet) and a text appearance adaptation network (TAANet). TLPNet first predicts the suitable regions for text embedding, after which TAANet adaptively adjusts the geometry and color of the text instance to match the background context. After training, those networks can be integrated and utilized to generate the synthetic dataset for scene text analysis tasks. Comprehensive experiments were conducted to validate the effectiveness of the proposed LBTS along with existing methods, and the experimental results indicate the proposed LBTS can generate better pretraining data for scene text detectors. Our dataset and code are made available at: \href{https://github.com/iiclab/DecompST}{https://github.com/iiclab/DecompST}.
\end{abstract}

\begin{IEEEkeywords}
Scene text synthesis, data augmentation, scene-text detection.
\end{IEEEkeywords}

\IEEEpeerreviewmaketitle

\section{Introduction}

\IEEEPARstart{D}{eep} neural networks have demonstrated remarkable success in the field of scene text detection and recognition, yet their performance heavily depends on the quantity and quality of the labeled training data. However, manual collection and labeling of images are costly in terms of both time and resources, and automatic data generation is expected. The image synthesis technique that composes text instances on background images offers a cost-effective and scalable alternative to manual annotation, and this approach has attracted increasing interest in the computer vision community. 

Various approaches have been investigated in the development of generation engines for synthetic scene-text images. Initially, based on the observation of real-world data, a set of sophisticated rules has been proposed to guide the design of generation engines. Gupta \textit{et al.} \cite{gupta_synthetic_2016} and Zhan \textit{et al.} \cite{ZhanVISD2018} generated synthetic text images from two-dimensional (2D) background images based on different strategies such as region selection, text warping, and text color matching. Liao \textit{et al.} \cite{LiaoST3D2020} and Long \textit{et al.} \cite{LongUnreal2020} further proposed rendering text on the surface of models in three-dimensional (3D) virtual worlds using Unreal Engine. Although realistic occlusions, perspectives, and illuminations can be realized in 3D engines, there is still a gap between the virtual and real worlds. To eliminate heuristic rules and complex setups, Yang \textit{et al.} \cite{Yanglearnbased2020} proposed a learning-based method consisting of a location module and an appearance module. The location module employs a conditional variational auto-encoder (cVAE) \cite{SohnCVAE2015} to learn the distribution of text locations directly from the original scene-text image and corresponding text bounding boxes (BBoxes). During training, the cVAE takes a scene text image as input, while during inference, a pure background image is used as input. The ``condition" is changed during the training and inference process, which is unreasonable and may limit its performance.

\begin{figure}[!t]
\centering
\includegraphics[width=\linewidth]{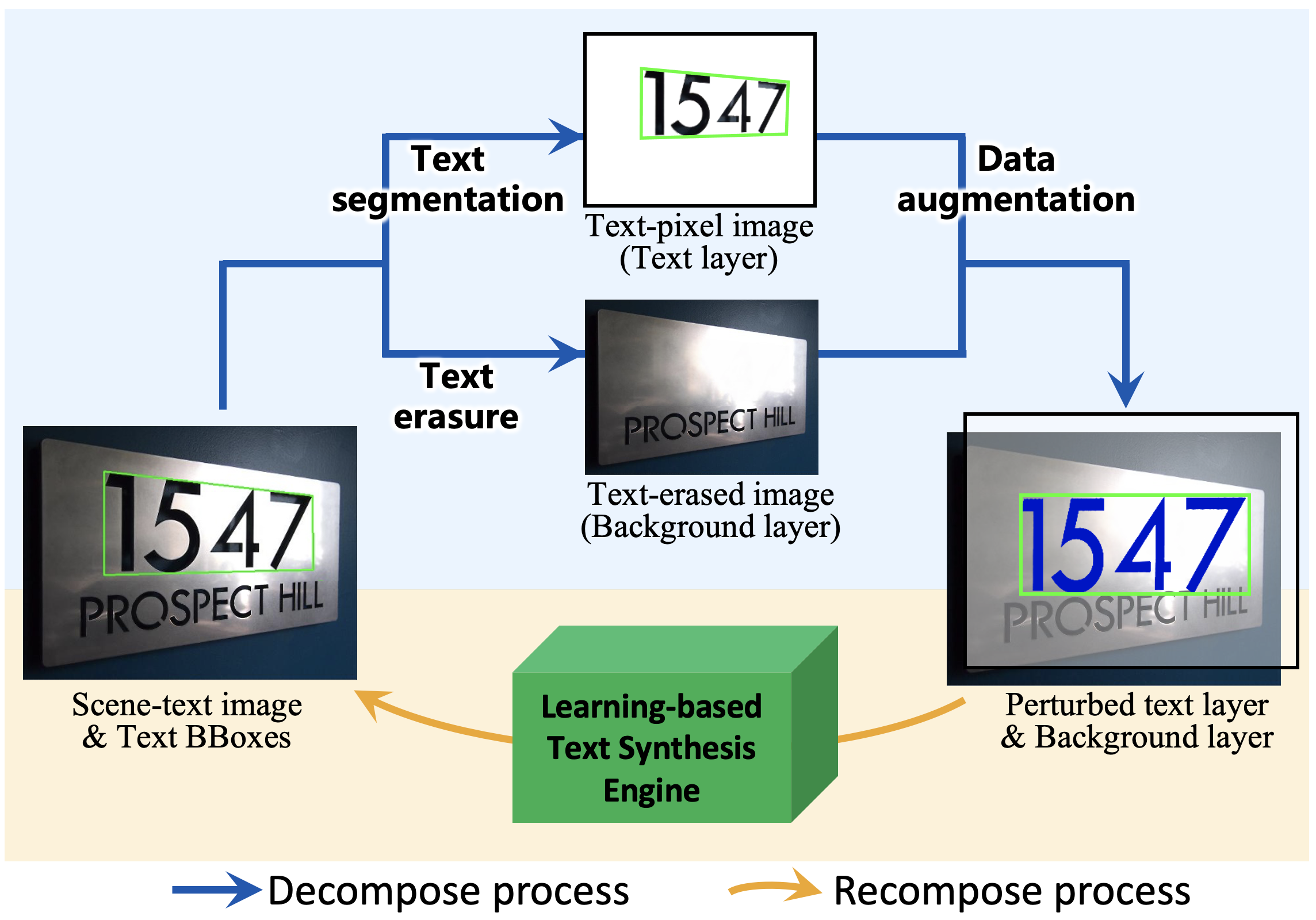}
\caption{Concept of our proposal. We first decomposed the real-world scene-text image into a text layer and a background layer. Next, we applied data augmentation to perturb the geometry and color of the text layer. Then, we proposed a Learning-based Text Synthesis Engine to recompose the two layers back to their original natural relationship, so that the engine can learn the complicated layout and appearance of text instances from real-world scene images}
\label{fig1.1}
\end{figure}

In this study, we aim to address the challenge of inadequate training data and facilitate learning-based text synthesis methods. To this end, we propose the DecompST dataset, which enables the decomposition of real-world scene text images into pure background images and pure text instances. These decomposed data can be utilized to train robust neural networks to learn the complicated layout and appearance of text instances in real-world scene images. The overall concept is illustrated in Fig.~\ref{fig1.1}.
Building upon the DecompST dataset, we propose a Learning-Based Text Synthesis engine (LBTS) that mainly includes a text location proposal network (TLPNet) and text appearance adaptation network (TAANet). TLPNet first predicts suitable regions from the background images for text embedding. TAANet then adaptively changes the perspective and color of the synthetic text instance to match the background. Once the networks have been effectively trained, an integrated data generation pipeline can be built to produce a scalable volume of synthetic data, which can subsequently be utilized as training data for various scene-text analysis tasks.

The main contributions of our study are summarized as follows:
\begin{itemize}
\item[\textbullet ] We introduce the DecompST dataset, which is able to decompose real-world scene-text images into separate pure background images and text instances, for the training of learning-based scene-text synthesis methods.

\item[\textbullet ] We propose a learning-based scene-text image synthesis engine (LBTS) that consists of a text location proposal network and a text appearance adaptation network, to generate realistic synthetic scene-text images.

\item[\textbullet ] The quality of our generated dataset and other existing synthetic datasets is evaluated by the performance of a baseline text detector. The experimental results demonstrate that our method can generate better pretraining data for scene text detectors than other state-of-the-art methods.

\end{itemize}

The structure of this paper is organized as follows. Section \ref{Related work} reviews related studies on scene-text detection, image synthesis, and data augmentation for scene-text analysis. Section \ref{Dataset} provides details about the proposed DecompST dataset. Section \ref{Methodology} introduces the proposed method, including the flow of data preprocessing and the structure of the two networks. In Section \ref{Experiment}, we evaluate and compare our proposed method with related synthetic datasets based on experimental results. Finally, concluding statements are presented in Section \ref{Conclusion}.

\section{Related work} 
\label{Related work}
\subsection{Scene Text Detection} 

With the rise of deep learning, scene text detection has been dramatically reshaped and facilitated, showing promising performance compared to traditional manual feature engineering algorithms \cite{chen_textnaturalscenes_2004,Neumann_textlocandrec_2011,jamil_edge-based_2011,mosleh_automatic_2013,huang_textloc_2013}. Recent learning-based scene text detection methods have been inspired by general object detection and image segmentation methods, which can be roughly categorized into regression-based and segmentation-based methods.
Regression-based methods aim to predict the bounding boxes of text instances directly. 
TextBoxes \cite{Liaotextbox2017} modified the anchors in the SSD \cite{liu_ssd_2016} to handle text with various aspect ratios. 
CTPN \cite{tian_CTPN_2016} combines the framework of Faster R-CNN \cite{RenfasterRCNN2015} with a recurrence mechanism to predict the contextual and dense fixed-width proposals of text. 
RRPN \cite{ma_RRPN_2018} proposes a rotation region proposal based on Faster R-CNN to bind arbitrary-oriented text with rotated rectangles. 
EAST \cite{zhou_east_2017} proposes a simplified detection pipeline that directly regresses rotated rectangles or quadrangles of text without using anchors. 
LOMO \cite{zhang_look_2019} improved the performance of EAST on the long text and arbitrarily shaped scene text by iteratively refining the preliminary proposals and considering the geometric properties of scene text. 

Segmentation-based methods usually first extract text from the segmentation map and then compute the text bounding boxes by post-processing. Zhang \textit{et al.} \cite{ZhangMutidetectFCN2016} integrated semantic labeling using FCN and MSER for pixel-level multi-oriented text detection. The Mask textspotter \cite{Lyumasktextspotter2018} was inspired by the framework of Mask R-CNN \cite{he_maskRCNN_2017} and performed character-level instance segmentation for each alphabet; thus, it has the ability to detect and recognize irregular text. TextSnake \cite{long_textsnake_2018} proposed a novel and flexible representation of arbitrarily shaped text and predicted heat maps of text centerlines, text regions, radii, and orientations to extract text instances. PSENet \cite{LiPSENet2020} gradually expanded small text kernels to complete shapes using multiple segmentation maps to effectively split close text instances. Liao \textit{et al.} \cite{LiaoDB2020} proposed a differentiable binarization (DB) module in a simple segmentation network to perform binarization. CRAFT \cite{BaekCRAFT2019} exploited the affinity between characters in the form of a heat map and proposed a weakly supervised framework to estimate character-level ground truths in existing real word-level datasets. ACE \cite{DaiACE2022} proposed to evolve the key points of the horizontal bounding box towards the corner points to detect arbitrarily-oriented objects or text.

\subsection{Image Synthesis} 
Inserting foreground objects into a background image is one of the most common image synthesis approaches for generating a photo-realistic composite image, which may face inconsistency problems between the foreground and background in the geometry and appearance domains. To solve these inconsistency problems, many subtasks have been investigated, such as object placement, image blending, image harmonization, and shadow generation.
Before the deep-learning era, many researchers explored automated image blending and harmonization. These methods transfer the color from one image to another based on the low-level statistics of the images, such as color distribution or histograms \cite{Reinhard2001,Cohen-Or2006,Xuehist2012}, gradient-domain information \cite{Perezpoisson2003,Jiagrad2006,darabi_inpaintpatch_2012,Taograd2013}, and multi-scale statistical features \cite{Sunkavallimulti2010}, among others.

With the emergence of neural networks, more challenging tasks have been investigated. 
ST-GAN \cite{LinSTGAN2018} seeks the geometric realism of image compositing by integrating a generative adversarial network (GAN) and spatial transformer networks (STNs) \cite{JaderbergSTN2015} to warp the foreground object in an iterative fashion. 
SF-GAN \cite{ZhanSFGAN2019} combines an STN and CycleGAN \cite{ZhuCycleGAN2017} to perform geometry transformation and appearance domain translation concurrently with an end-to-end trainable network. Benefiting from the designed structure, the SF-GAN can also achieve synthesis realism in both geometry and appearance spaces without using paired training data.
GCC-GAN \cite{ChenGCC2019} was proposed to address geometric and color consistency in composite images by integrating four subnetworks: a transformation network, a refinement network, a discriminator network, and a segmentation network. In the transformation network, not only are the parameters of the transformation matrix predicted, but the parameters of linear color transformation that control the contrast and brightness are also predicted simultaneously.
Tsai \textit{et al.} \cite{Tsaideepharmo2017} introduced an end-to-end image harmonization network with a shared encoder and two decoders, where the learned semantic information was used to facilitate harmonization.
Inspired by AdaIN \cite{HuangAdaIN2017}, Ling \textit{et al.} \cite{LingRAIN2021} treated image harmonization as a background-to-foreground style transfer problem and proposed a plug-and-play region-aware adaptive instance normalization (RAIN) module that explicitly formulates the visual style from the background and adaptively applies it to the foreground.

\subsection{Data Augmentation for Scene Text Analysis} 
The text synthesis technique, which involves inserting text instances into scene background images, was initially investigated as a data augmentation approach for the training of scene text detection and recognition models. Later, synthetic datasets were utilized as important training data for other tasks such as scene text segmentation \cite{QintextsegWACV2018,Bonechicocoseg2020}, scene text erasing \cite{TursunMTRNet2019,Tangstroke2021}, and scene text editing \cite{WuSRNet2019,Yangswaptext2020}. 

Wang \textit{et al.} \cite{Wangtextrecog2012} generated a character-centered synthetic image to train a character-level scene-text recognition model. Jaderberg \textit{et al.} \cite{JaderbergMJ2016} generated a word-centered synthetic dataset using a set of predefined random processes, including font selection and rendering, bordering/shadowing and coloring, layer composition, projective distortion, blending, and noise addition. SF-GAN \cite{ZhanSFGAN2019} was trained without paired data because of its unsupervised pipeline, which can also be applied in text synthesis tasks to generate patch-level synthetic text images. Yim \textit{et al.} \cite{Yimtiger2021} further analyzed existing synthesis techniques \cite{JaderbergMJ2016,gupta_synthetic_2016} and integrated the effective parts as a new-generation engine for scene text recognition tasks. These methods generate text-centered images, whose applications are limited. 

Gupta \textit{et al.} \cite{gupta_synthetic_2016} first attempted to synthesize text in the wild to generate the SynthText dataset, which is beneficial for training scene-text detection tasks. The SynthText engine finds suitable text embedding regions in the background image following a set of rules that consider semantic segmentation maps and depth maps, and it renders text instances with color selection, perspective distortion, and Poisson blending \cite{Perezpoisson2003} according to the local background information. 
Zhan \textit{et al.} \cite{ZhanVISD2018} exploited saliency-guided ``semantic coherent" image synthesis by leveraging the annotations of semantic segmentation map and visual saliency map. They also designed an adaptive text appearance mechanism to determine the color and brightness of texts by matching a list of pairs, which includes the HoG feature of the background and LAB space statistics of text, gathered from real scene-text images. 
Yang \textit{et al.} \cite{Yanglearnbased2020} proposed a learning-based, data-driven text synthesis engine by dividing the text synthesis into two sub-tasks:1) determining the location of text and 2) making the appearance of the inserted text more realistic. A conditional variational auto-encoder \cite{KingmaVAE2014, SohnCVAE2015} was utilized to learn the distribution of text locations from real-world data, and a masked Cycle-GAN \cite{ZhuCycleGAN2017} was proposed to translate the appearance of synthetic images to the real-data domain.
In contrast to rendering text in 2D static images, Long \textit{et al.} \cite{LiaoST3D2020, LongUnreal2020} renders text and the scene as integrity in 3D virtual worlds using the Unreal Engine.  In this way, real-world variations, including complex yet correct perspective distortions, various lighting conditions, and occlusions, can be realized in the synthesized scene text images.

\begin{figure}[!t]
\includegraphics[width=\linewidth]{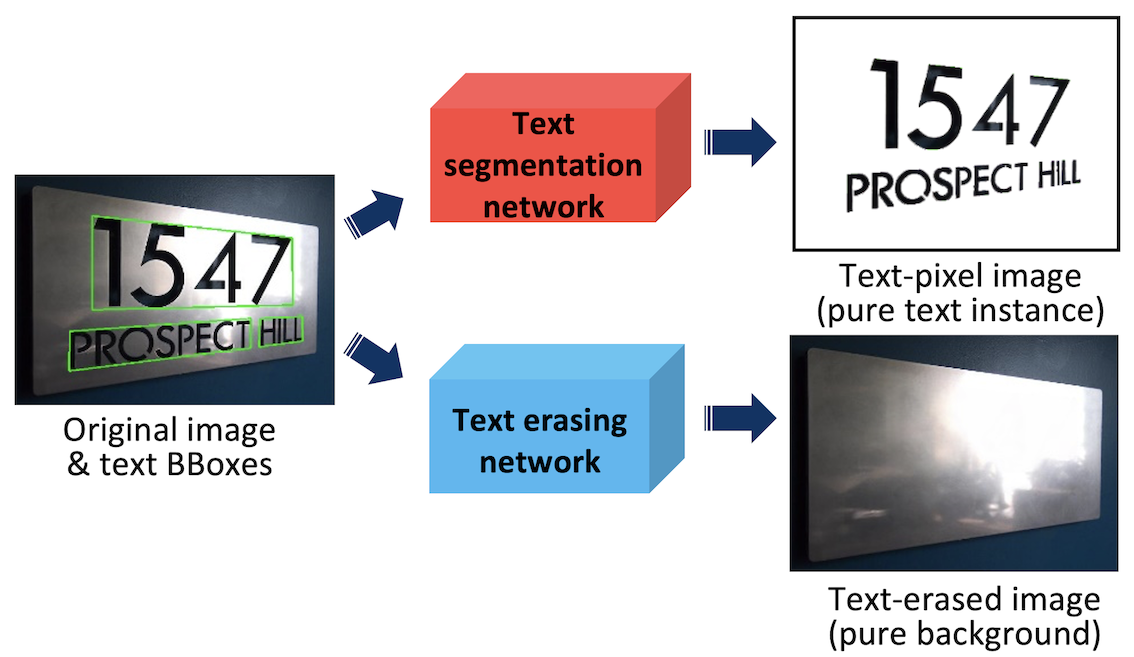}
\caption{Given the original image and corresponding text BBoxes, we decompose real-world scene-text images into pure background images and text instances using a text-erased image and stroke-level text mask.}
\label{fig3.0}
\end{figure}

In terms of learning-based methods for synthesizing scene-text images, our method is closely related to the method proposed in \cite{Yanglearnbased2020}. Their approach samples latent vectors from the prior distribution and feeds them to a cVAE to directly output the affine transformation parameters, which are used to globally transform the location and perspective of text instances. However, owing to the direct use of scene text images and the corresponding text BBoxes for training, the ``condition'' of cVAE is changed during the training and inference processes, which may achieve unsatisfactory performance. Our proposed DecompST dataset can address this problem by providing a data pair of text-erased images and original text BBoxes.

Another closely related method is presented in \cite{ZhanSFGAN2019}, which can concurrently achieve realism in both geometry and appearance spaces without supervision by employing an innovative network structure. In addition, the method in this study can generate patch-level synthetic text images for scene-text recognition tasks. In contrast to their work, our proposed method is a fully supervised image synthesis method that leverages the DecompST dataset, aiming to train more robust networks to generate image-level synthetic scene-text images specifically for the text detection task.

\section{DecompST Dataset}
\label{Dataset}
We introduce a dataset called DecompST, which is a quadruplet of the original scene-text images, text BBoxes, text-erased images, and stroke-level text masks. This dataset can decompose real-world scene-text images into pure background images and text instances, as shown in Fig.~\ref{fig3.0}. Those components can be utilized to train a robust network to learn the complicated layout and appearance of text instances in real-world scene images. We have made this dataset publicly available and hope that it can motivate more learning-based scene text synthesis methods to generate high-quality synthetic training data for scene text detection and recognition tasks.

\begin{figure}[!t]
\setlength{\fboxrule}{0.4pt}
\setlength{\fboxsep}{0pt}
\hspace{-7pt}
\subfigure[]{
\begin{minipage}[t]{0.325\linewidth}
    \includegraphics[height=0.9\textwidth, width=1\textwidth]{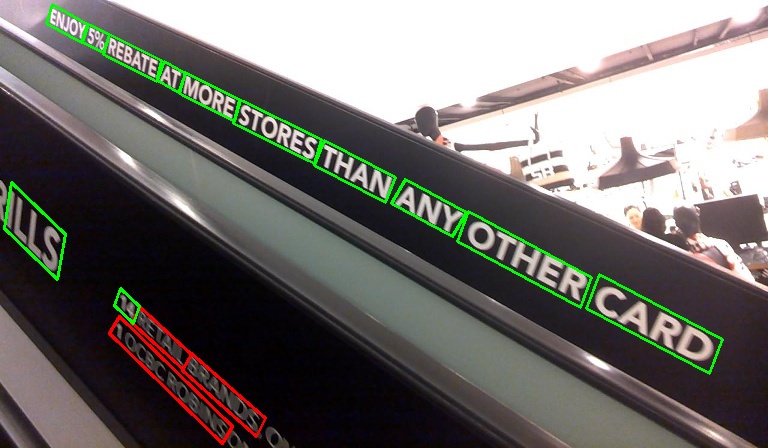}\vspace{2pt}
    \includegraphics[height=0.9\textwidth, width=1\textwidth]{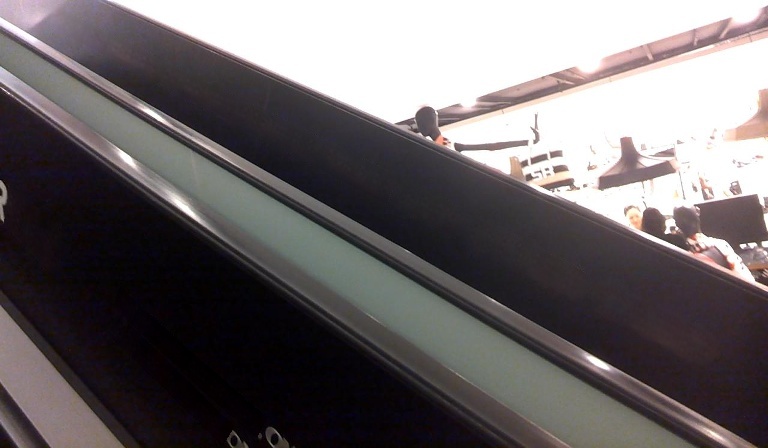}\vspace{2pt}
    \includegraphics[height=0.9\textwidth, width=1\textwidth]{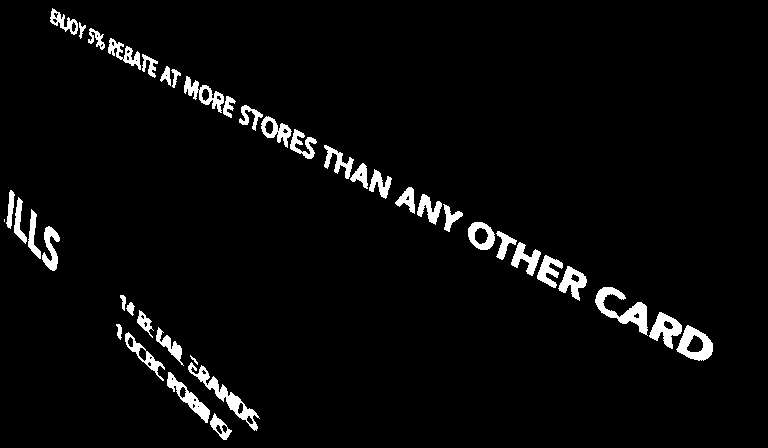}\vspace{2pt}
    \fbox{\includegraphics[height=0.9\textwidth, width=1\textwidth]{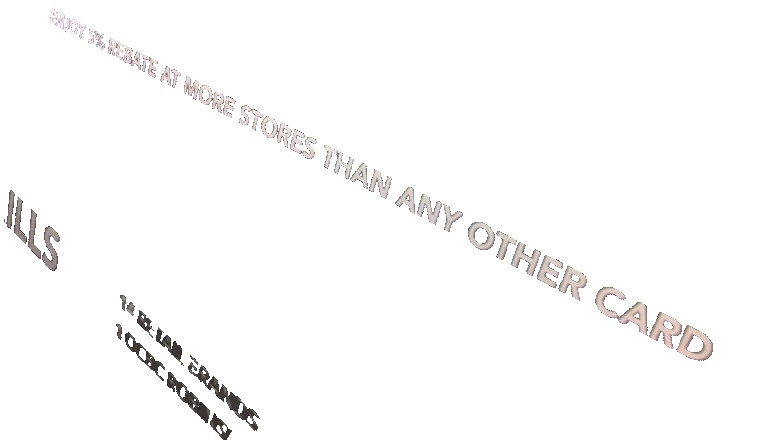}}\vspace{4pt}
\end{minipage}}
\hspace*{-8.5pt}
\subfigure[]{
\begin{minipage}[t]{0.325\linewidth}
    \includegraphics[height=0.9\textwidth, width=1\textwidth, trim=0 0 70 0, clip ]{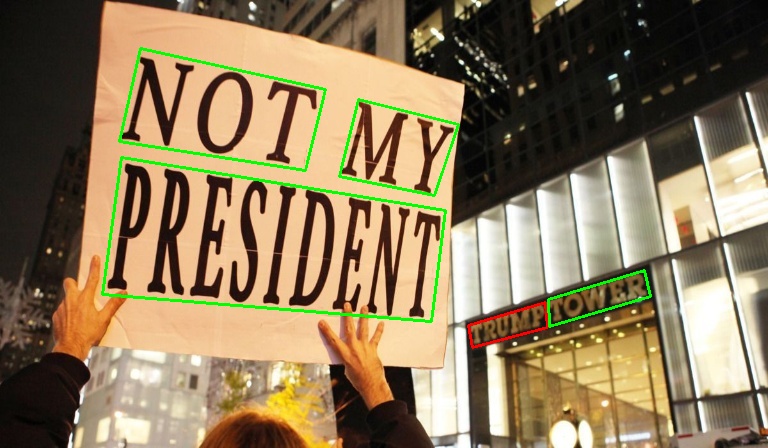}\vspace{2pt}
    \includegraphics[height=0.9\textwidth, width=1\textwidth, trim=0 0 70 00, clip ]{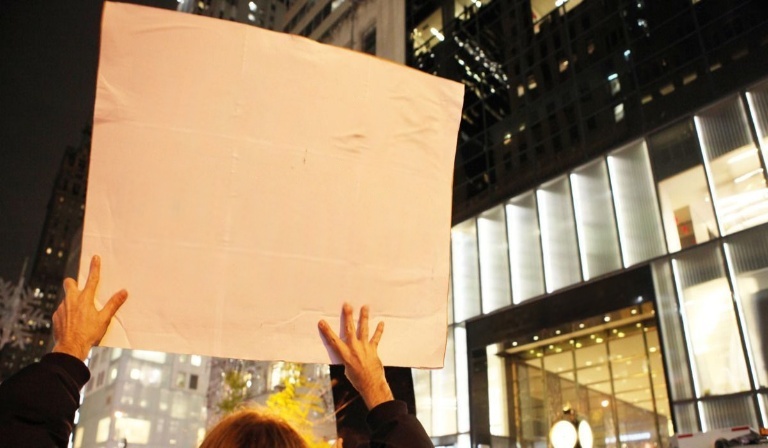}\vspace{2pt}
    \includegraphics[height=0.9\textwidth, width=1\textwidth, trim=0 0 70 00, clip ]{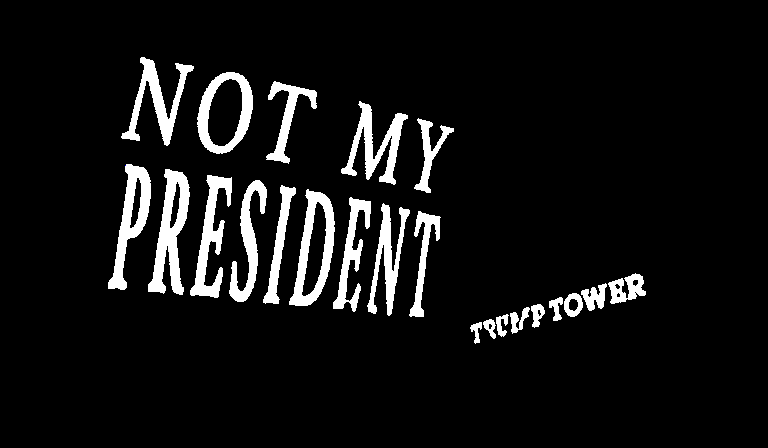}\vspace{2pt}
    \hspace{-1pt}\fbox{\includegraphics[height=0.9\textwidth, width=1\textwidth, trim=0 0 70 0, clip ]{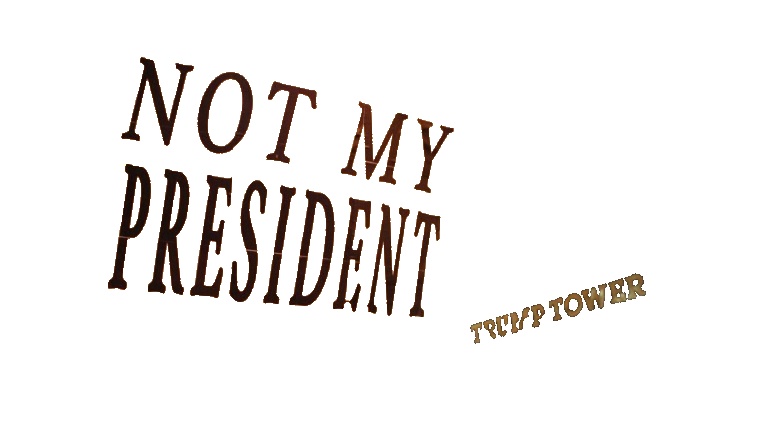}}\vspace{4pt}
\end{minipage}}
\hspace*{-8.5pt}
\subfigure[]{
\begin{minipage}[t]{0.325\linewidth}
    \centering
    \includegraphics[height=0.9\textwidth, width=1\textwidth, trim=0 110 0 90, clip ]{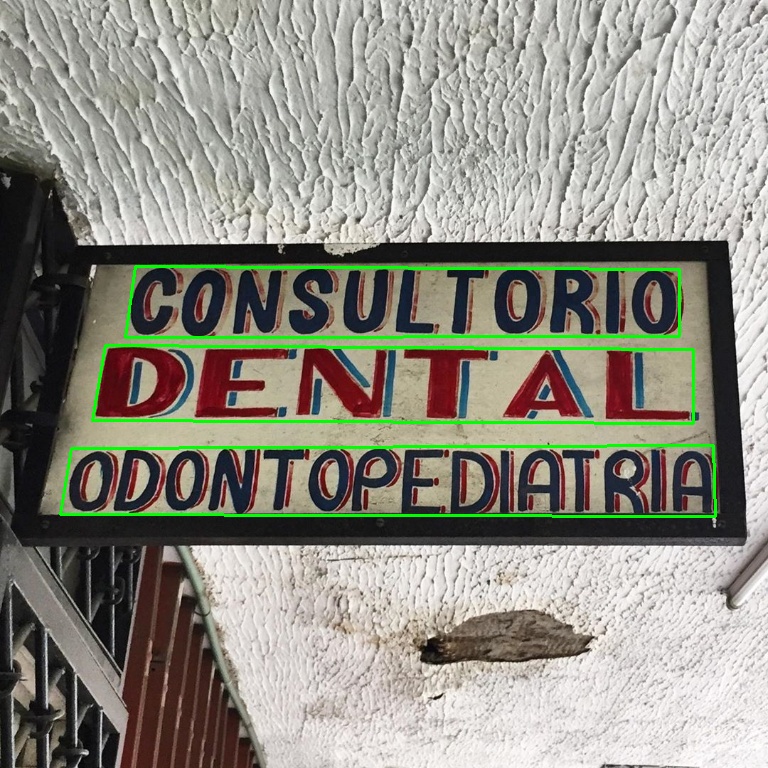}\vspace{2pt}
    \includegraphics[height=0.9\textwidth, width=1\textwidth, trim=0 110 0 90, clip ]{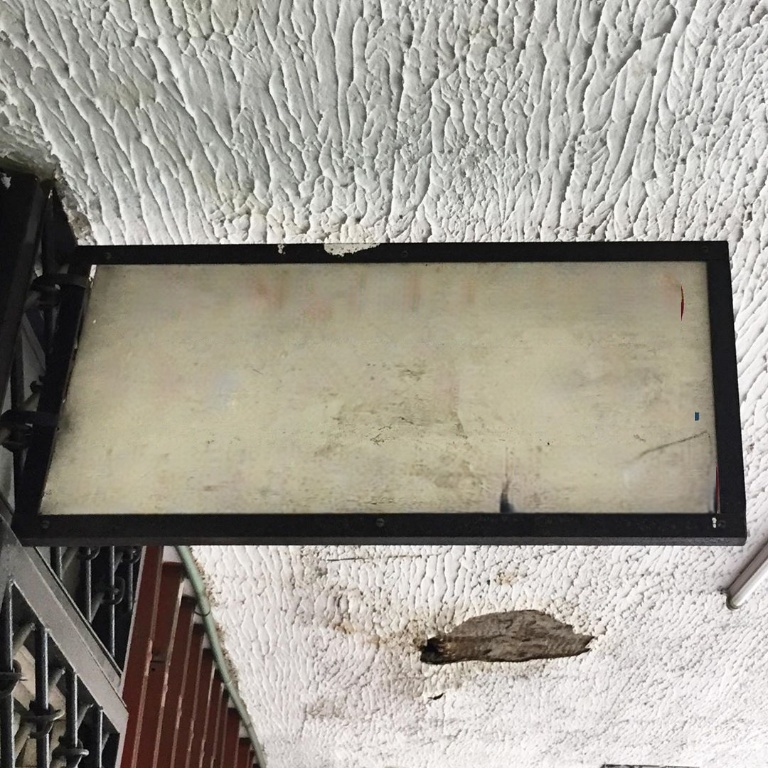}\vspace{2pt}
    \includegraphics[height=0.9\textwidth, width=1\textwidth, trim=0 110 0 90, clip ]{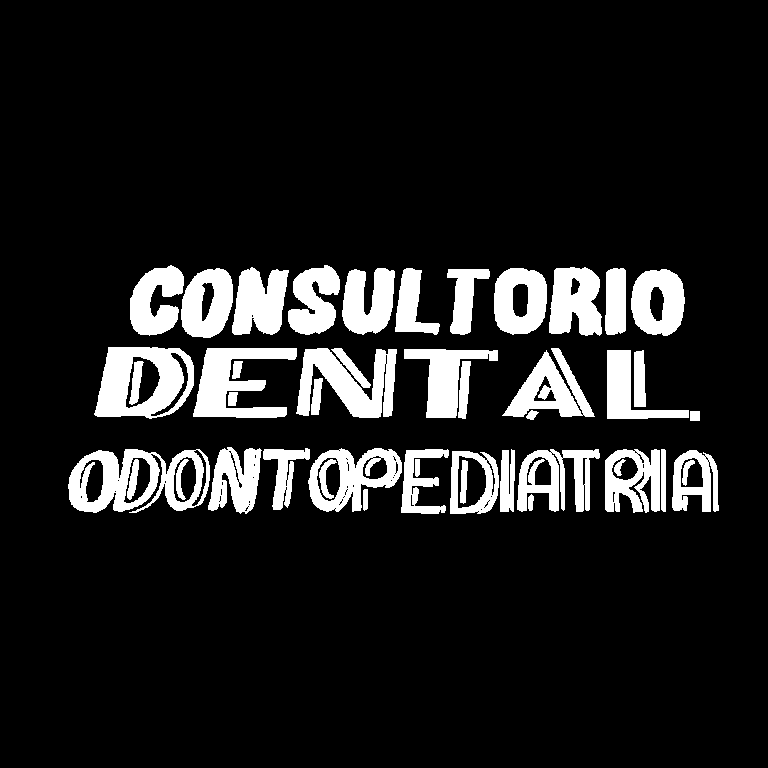}\vspace{2pt}
    \fbox{\includegraphics[height=0.9\textwidth, width=1\textwidth, trim=0 110 0 90, clip ]{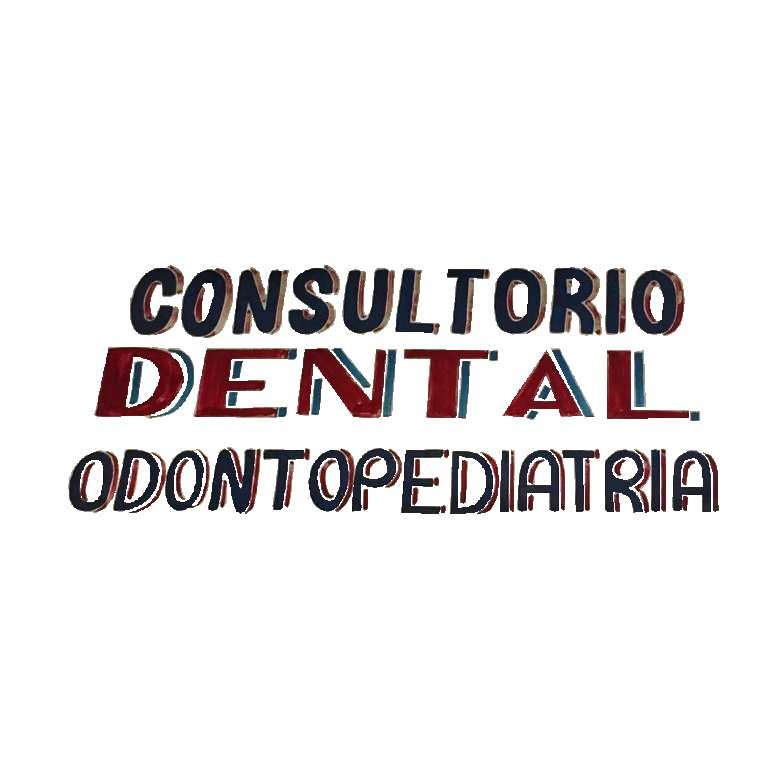}}\vspace{4pt}
\end{minipage}}
\hspace*{-8.5pt}
\caption{Some image samples from our proposed DecompST Dataset. The first row contains the original images with text BBoxes, where valid text instances are marked in \textcolor{green}{green} BBoxes and invalid ones are in \textcolor{red}{red} BBoxes. The second row is our generated text-erased images. The third row is the stroke-level text masks. The fourth row is the text-pixel images masked by stroke-level masks.} \label{fig3.1}
\end{figure}

\begin{table}[t]
\centering
\caption{Number of Images and Valid Text Instances from Different Source Datasets.}
\label{tab3.2}
\begin{tabular}{ccc}
\toprule
                & Images & Text Instances \\ \midrule
IC15 \cite{KaratzasICDAR2015}       & 787    & 1848           \\
MLT19  \cite{NayefICDAR2019MLT}      & 1681   & 6652           \\
SegText \cite{XuTextSeg2021}        & 2117   & 7517           \\ \midrule
Total                                & 4585   & 16017          \\ \bottomrule
\end{tabular}
\end{table}

\subsection{Image Collection} 
All the images in our dataset were collected from several public real-world scene text detection benchmarks, including the ICDAR-2015 \cite{KaratzasICDAR2015}, MLT-2019 \cite{NayefICDAR2019MLT}, and TextSeg \cite{XuTextSeg2021} datasets. The ICDAR-2015 \cite{KaratzasICDAR2015} and MLT-2019 \cite{NayefICDAR2019MLT} datasets are classic benchmarks for scene text detection. The TextSeg \cite{XuTextSeg2021} dataset, on the other hand, specifically focuses on scene text segmentation. It provides comprehensive annotations encompassing quadrilateral BBoxes at both word and character levels, along with pixel-level text masks. We opted to use the TextSeg dataset because its manually-labeled, high-quality pixel-level text masks align with our requirements for stroke-level text masks.
For each dataset, we collected both the training and validation sets, but we only selected Latin and Chinese parts of the MLT-2019 \cite{NayefICDAR2019MLT} dataset, and the scene-image part of the TextSeg \cite{XuTextSeg2021} dataset.

\subsection{Annotation Details} 
This section provides a detailed description of the annotation process applied to create the DecompST dataset. For each text instance in the collected images, our goal was to obtain the corresponding text-erased patch and stroke-level text mask. Since the text instances in images are already labeled by BBoxes, we utilized a word-level scene-text-erasing method \cite{Tangstroke2021} to erase each text instance individually and generate text-erased images. 
To obtain the stroke-level text mask of the ICDAR-2015 \cite{KaratzasICDAR2015} and MLT-2019 \cite{NayefICDAR2019MLT} datasets, we employed the stroke mask prediction module (SMPM) in \cite{Tangstroke2021} to extract the pixel-level text mask. However, as the original SMPM was designed to  predict a dilated text mask, we retrained the SMPM using the same synthetic dataset \cite{Tangstroke2021}, but with original-size text masks as ground truth. Subsequently, this retrained SMPM was utilized to accurately predict text masks that precisely fit the text instances.
Given that predictions made by neural networks can sometimes be imperfect, it is necessary to manually label the quality of predicted results. 

Our labeling criteria for text-pixel images focused on the readability of text and the integrity of the text mask. As for text-erased images, we assessed the quality based on the effectiveness of text erasure and the restoration of the background. 
During the annotation process, the annotators checked the text-pixel image and text-erased image of each text instance and labeled both their quality as 1 or 0, where 1 indicated good and 0 indicated bad. Only text instances that received 1 on both sides were considered valid data, and other data were deemed invalid.
For the TextSeg dataset, because accurate pixel-level text masks were provided, all text masks were labeled as 1, and we only assessed the quality of the text-erased image, assigning a label of 1 or 0.

Finally, the DecompST dataset contains 4585 images with 16017 valid text instances with corresponding text-erased images, stroke-level text masks, and quadrilateral bounding boxes, as summarized in Table~\ref{tab3.2}. Visual samples of annotated instances from the DecompST dataset are presented in Fig.~\ref{fig3.1}.


\begin{figure*}[!t]
\centering
\includegraphics[width=0.97\linewidth]{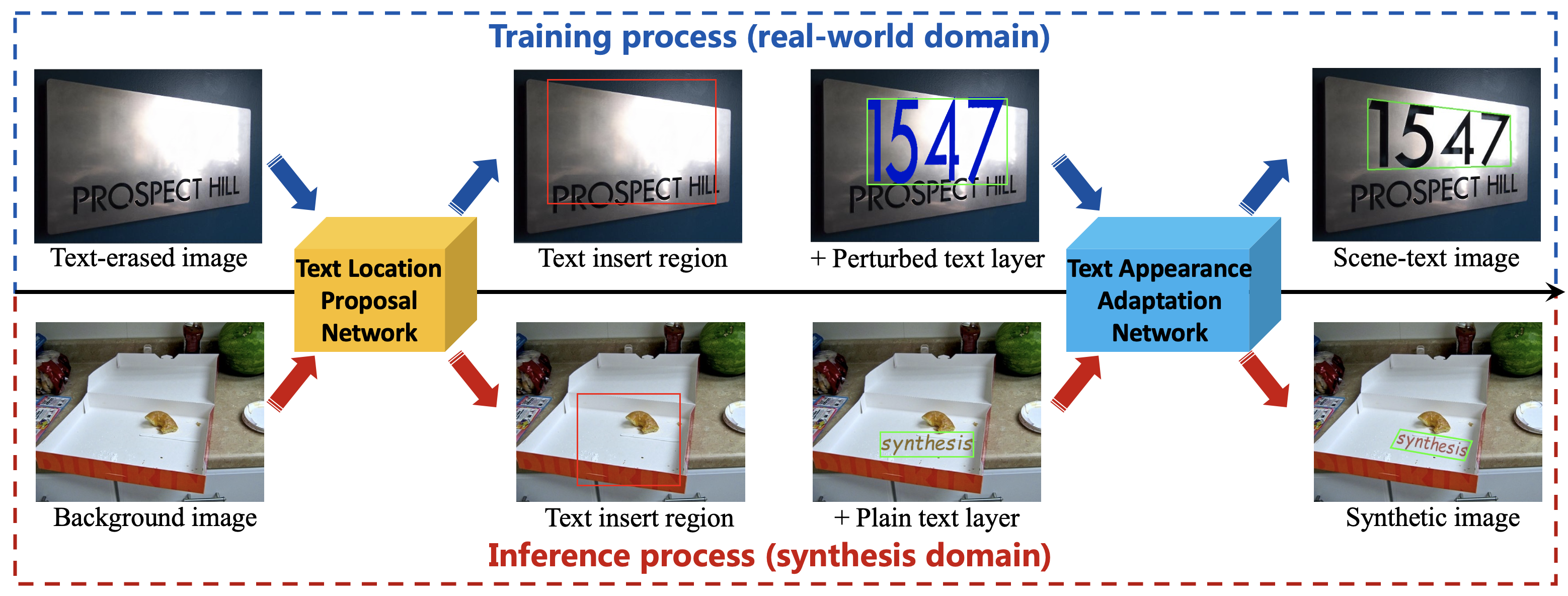}
\caption{Pipeline of our proposed Learning-based Text Synthesis Engine (LBTS). It mainly consists of two networks: text location proposal network (TLPNet) and text appearance adaptation network (TAANet). Given a background image, TLPNet predicts suitable regions for text embedding. Then, TAANet aligns the geometric and color relationship between the synthetic text instance and the background. We trained our proposed networks on decomposed real-world data and applied them in the synthesis domain to generate synthetic scene-text images.}
\label{fig4.0}
\end{figure*}

\section{Methodology} 
\label{Methodology}
In this section, we present our proposed learnable text synthesis (LBTS) method, which mainly consists of two subnetworks: the text location proposal network (TLPNet) and the text appearance adaptation network (TAANet), as illustrated in Fig.~\ref{fig4.0}. More concretely, during the training, given a text-erased image, TLPNet first predicts suitable regions for text embedding. Then, a perturbed text layer is added and TAANet adaptively adjusts the perspective and color of the perturbed text layer to restore its original natural appearance. After training, we can feed two networks with unseen background images and plain text images to generate synthetic scene-text images.
Further details regarding the network structure, training process, and inference strategy are presented in the following subsections.


\subsection{Text Location Proposal Network} 

\subsubsection{Data Preprocessing in training} 
\begin{figure*}[!t]
\centering
\subfigure[]{
\begin{minipage}[t]{0.196\linewidth}
    \includegraphics[width=1\textwidth]{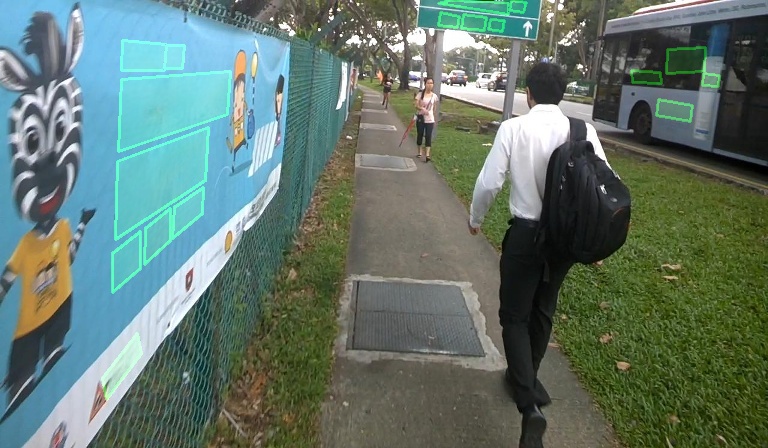}\vspace{2pt}
    \includegraphics[width=1\textwidth]{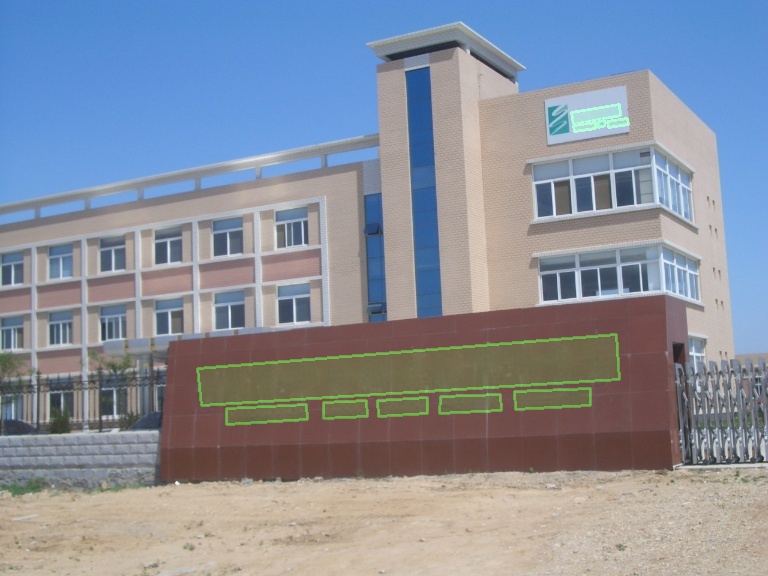}\vspace{2pt}
    \includegraphics[width=1\textwidth]{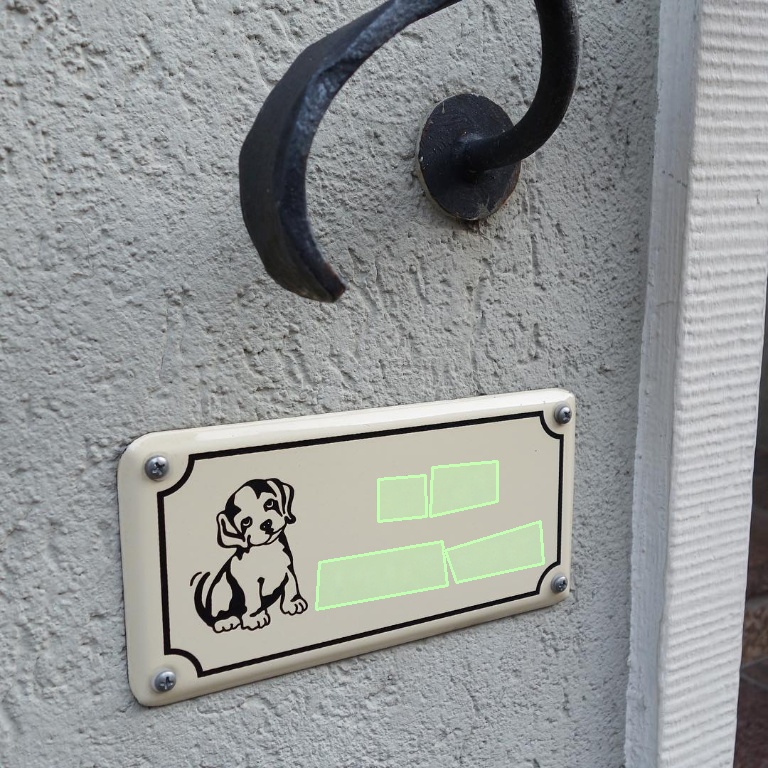}\vspace{4pt}
\end{minipage}}
\hspace*{-8.5pt}
\subfigure[]{
\begin{minipage}[t]{0.196\linewidth}
    \includegraphics[width=1\textwidth]{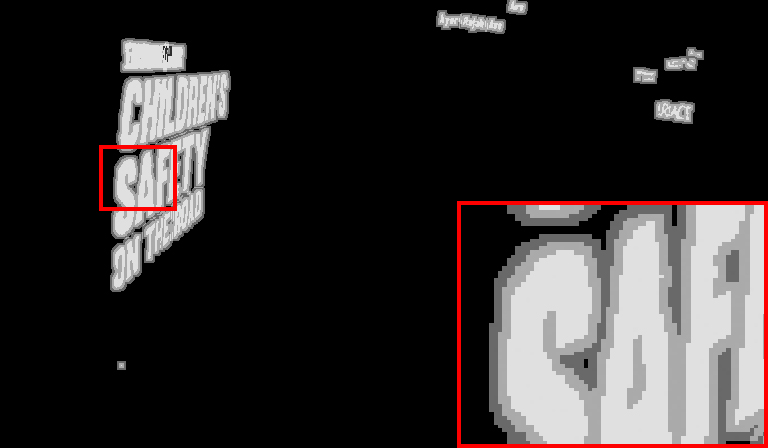}\vspace{2pt}
    \includegraphics[width=1\textwidth]{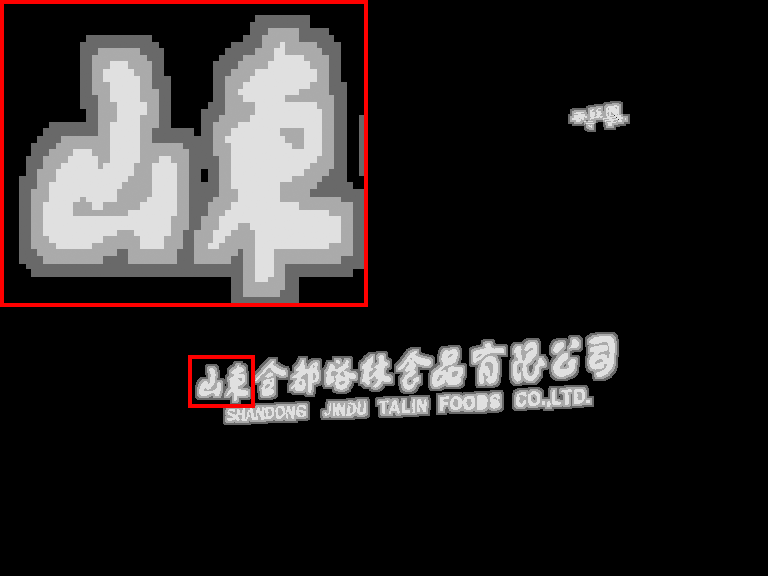}\vspace{2pt}
    \includegraphics[width=1\textwidth]{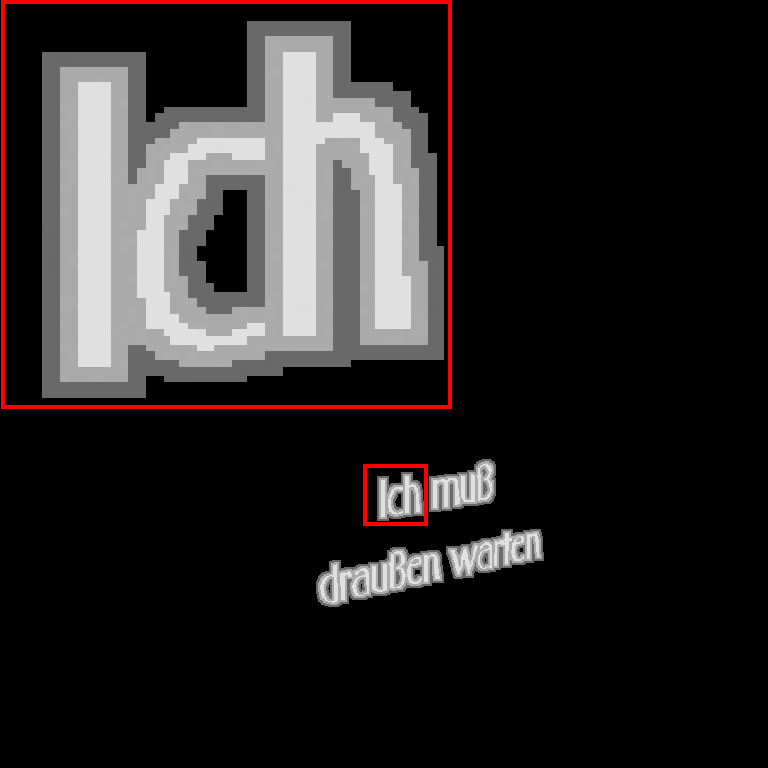}\vspace{4pt}
\end{minipage}}
\hspace*{-8.5pt}
\subfigure[]{
\begin{minipage}[t]{0.196\linewidth}
    \includegraphics[width=1\textwidth]{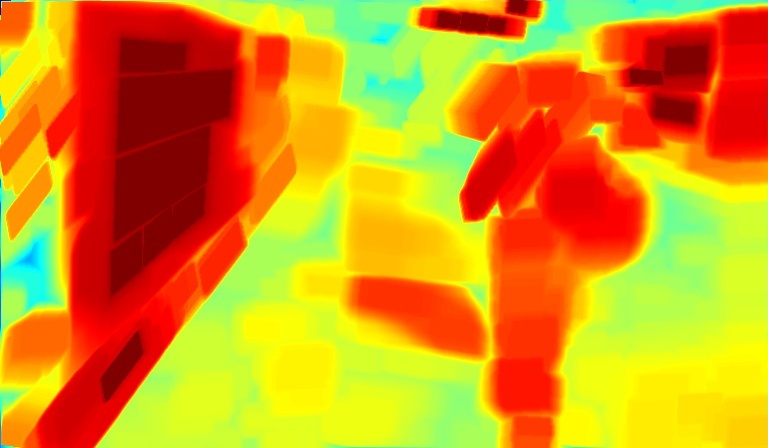}\vspace{2pt}
    \includegraphics[width=1\textwidth]{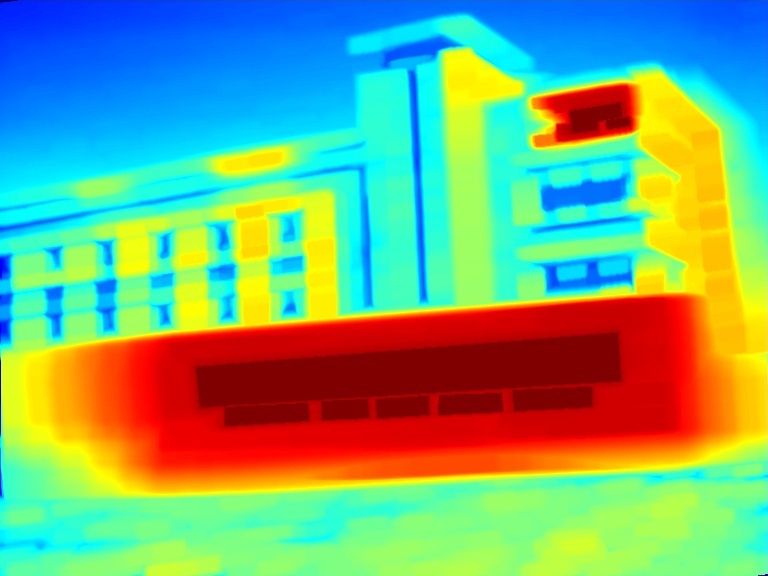}\vspace{2pt}
    \includegraphics[width=1\textwidth]{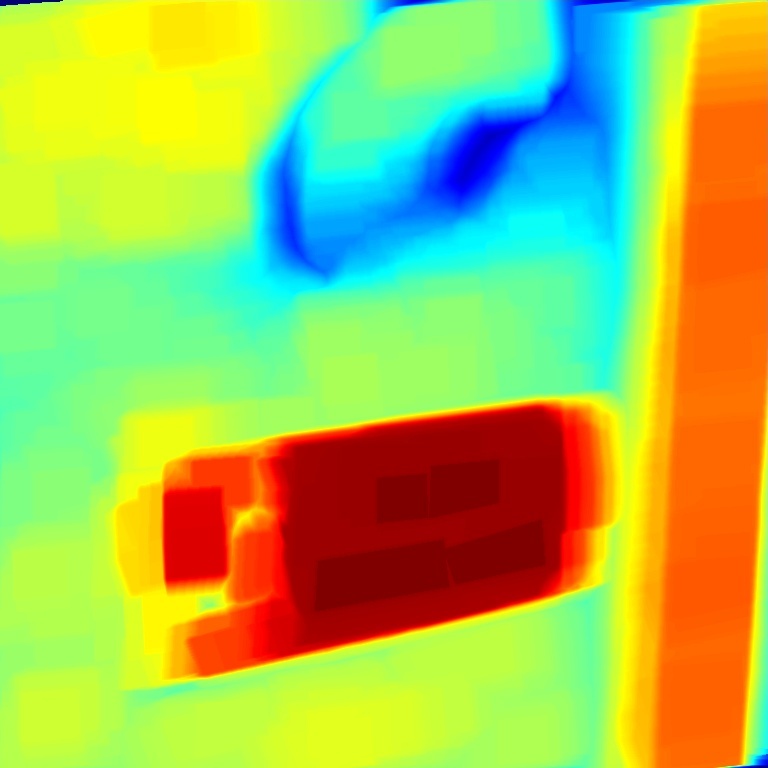}\vspace{4pt}
\end{minipage}}
\hspace*{-8.5pt}
\subfigure[]{
\begin{minipage}[t]{0.196\linewidth}
    \includegraphics[width=1\textwidth]{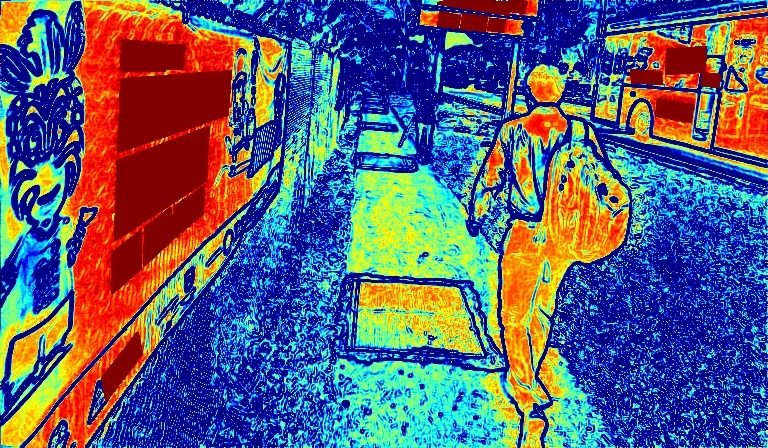}\vspace{2pt}
    \includegraphics[width=1\textwidth]{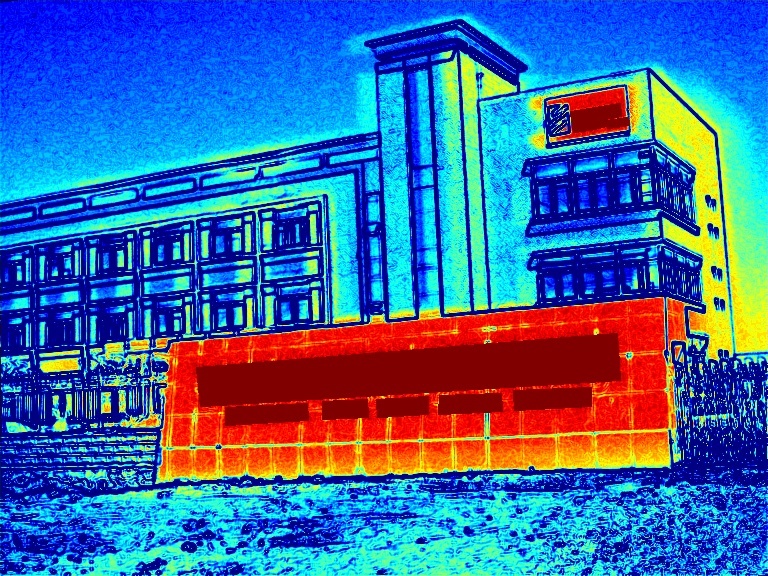}\vspace{2pt}
    \includegraphics[width=1\textwidth]{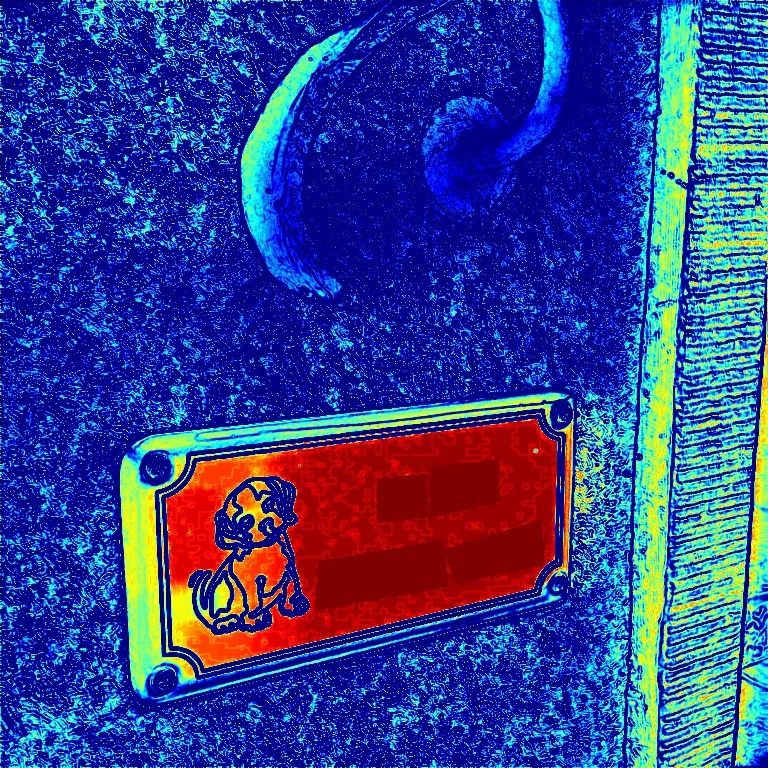}\vspace{4pt}
\end{minipage}}
\hspace*{-8.5pt}
\subfigure[]{
\begin{minipage}[t]{0.196\linewidth}
    \includegraphics[width=1\textwidth]{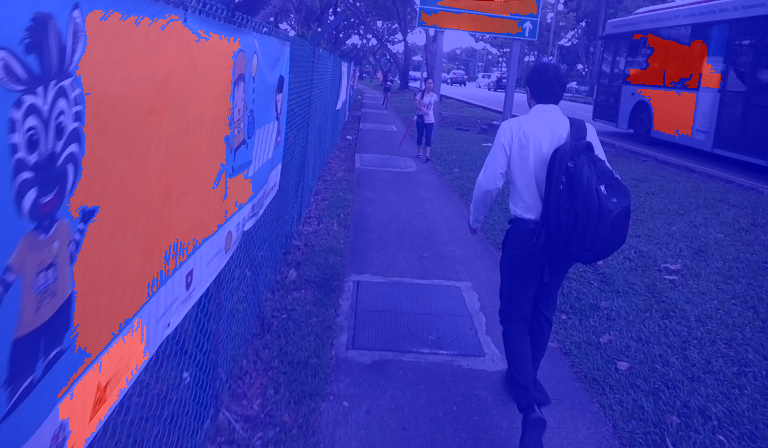}\vspace{2pt}
    \includegraphics[width=1\textwidth]{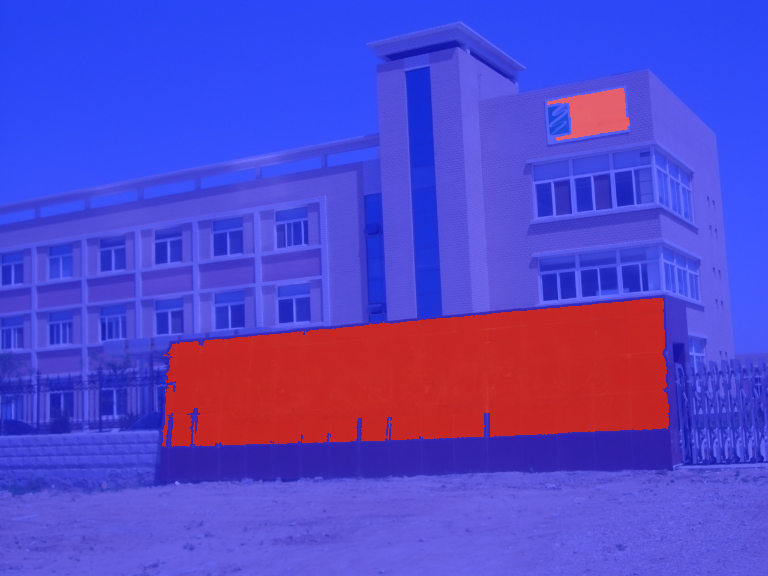}\vspace{2pt}
    \includegraphics[width=1\textwidth]{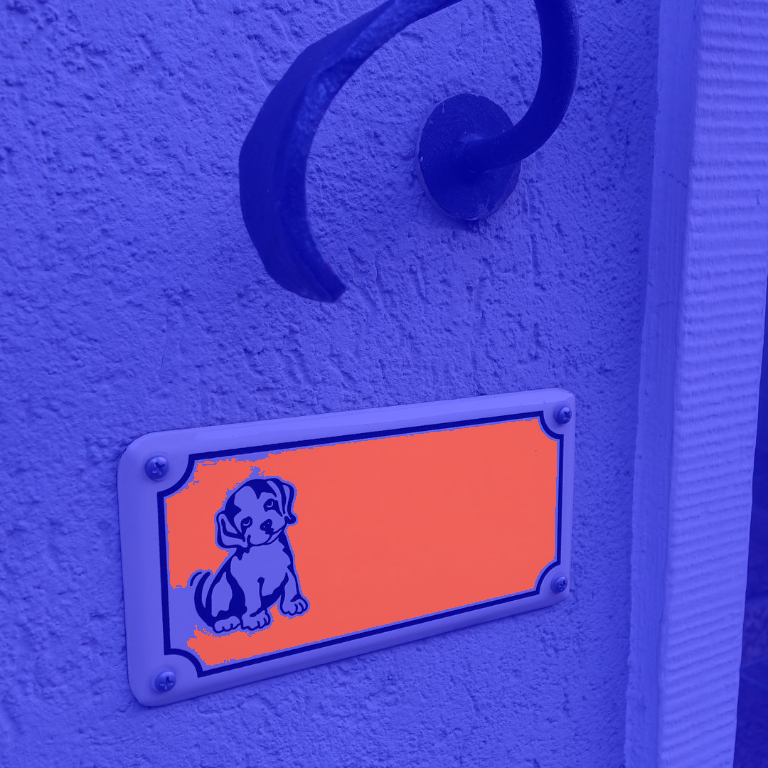}\vspace{4pt}
\end{minipage}}
\vspace{-4pt}
\caption{Flow of data preprocessing. (a) Text-erased image $I$ and $\it{BBOX}$ regions. (b) Visualization of $\mathcal{R}_{i}$, $w_{i}$. The $\mathcal{R}_{i}$ refer to the corresponding regions of each text instance and the brighter regions in (b) of $\mathcal{R}_{i}$ mean higher $w_{i}$. (c) Appearance consistency heatmap $H_a$. (d) Edge-based segmented heatmap $H_e$. (e) Final generated heatmap $H_f$ ($H_f$ is overlaid on $I$ for a better view). The red regions are treated as GT during the training of TLPNet. Note that the original $H_a, H_e, H_f$ are gray-scale images; we visualized them as heatmaps in this figure.} \label{fig4.1.0}
\end{figure*}

Undoubtedly, the regions within the original BBoxes can be regarded as the ground truth of the text region for learning. Furthermore, we consider that the feasible region for text embedding could be extended if the background shares a similar pattern in a neighboring area, especially in the case of scene text that usually appears in relatively plain regions, such as billboards, walls, and signs. 
To identify the regions that have a similar appearance to the text-erased regions, we adopted the concept of the appearance descriptor and appearance distance from InstaBoost \cite{Fanginstaboost2019} to measure the appearance similarity between text-erased regions and all other regions within an image. 
The appearance descriptor $\mathcal{D(\cdot)}$ is a combination of three weighted regions $\mathcal{R}_{i}$ of each valid text instance in the text-erased image, which is related to the corresponding text location:
\begin{equation}
\label{appearance descriptor}
\mathcal{D}(p_{x},p_{y}) = \{(\mathcal{R}_{i}(p_{x}, p_{y}), w_{i}) | i\in\{1,2,3\}\},
\end{equation}
where $\mathcal{R}_{1}$ denotes the region of the stroke-level mask, and $\mathcal{R}_{2}$ and $\mathcal{R}_{3}$ are the dilated contours of the stroke-level mask with different scales ($\mathcal{R}_{2}$ is the inner contour), given $p_{x}$, $p_{y}$ as the center of the instance. $w_{i}$ is the weight coefficient of $\mathcal{R}_{i}$, and $w_{1}>w_{2}>w_{3}$ is defined to emphasize the higher similarity around the inner neighboring areas of the original text instance. Fig.~\ref{fig4.1.0} (b) shows some examples of visualizations of the descriptor's region $\mathcal{R}_{i}$ and weight $w_{i}$. 

Next, given a target text appearance descriptor $\mathcal{D}_{t}(p_{tx},p_{ty})$, we assess the appearance similarity between the appearance descriptor of each pixel in the text-erased image and $\mathcal{D}_{t}$ using the appearance distance.
The appearance distance for a given pixel $(x, y)$, conditioned on $ \mathcal{D}_{t} $, can be formulated as follows:
\begin{equation}
\label{appearance distance}
d^{\mathcal{D}_t}_{(x,y)} =\hspace{-0.5cm} \min \limits_{(u, v)\in \it{BBOX}}{\sum\limits_{i=1}^3 \hspace{1.1cm} \smashoperator[lr]{ \sum_{\substack{(x_t,y_t) \in \mathcal{R}_{ti}(p_{tx},p_{ty}) \\(x_s,y_s) \in \mathcal{R}_{si}(p_{tx}-u+x,p_{ty}-v+y)}}}}  \hspace{0.3cm} w_{i} \Delta (I(x_t,y_t), I(x_s,y_s)),
\end{equation}
where $\it{BBOX}$ is the area inside the original text BBox. $I(x, y)$ denotes the RGB value of the text-erased image on $(x, y)$ pixel coordinates, and $\Delta$ is the Euclidean distance. 
The result of $\Delta$ is counted as infinity if $(x_s, y_s)$ is outside the boundary of the text-erased image.

By gathering the appearance distance of each pixel conditioned on the target text instance, we construct the target text appearance distance map $H_d^t$. $H(x, y)$ denote the value of the map $H$ at pixel coordinates (x, y). Consequently, $H_d^t(x, y) = d^{\mathcal{D}_{t}}_{(x,y)}$. We generate the corresponding appearance consistency heatmap $H_a^t$ by applying a normalization function to every pixel of the $H_d^t$, expressed as follows:
\begin{equation}
\label{Hat}
H_a^t(x, y) = \Big(1 - \frac {H_d^t(x,y)}{d_{\it max}}\Big)^3,
\end{equation}
here, $d_{max}$ is the maximum value in $H_d^t$ except the infinity. During the calculation of Eq.~\ref{Hat}, the infinity is set to $d_{max}$.

For each text instance in an image, we calculate the corresponding appearance consistency heatmaps and combine them into $H_a$:
\begin{equation}
\label{Ha}
H_a(x, y) = \max\limits_{k \in W}H_a^k(x, y),
\end{equation}
where $k \in W$ is the index of the text instance and $W$ denotes the set of valid text instances in the text-erased image.

Up to this point, the appearance consistency heatmap $H_a$ only takes into account the color similarity between patches of valid text instances and other patches in a text-erased image. Therefore, $H_a$ is redundant and lacks semantic information. To address this limitation, we propose a further processing method for $H_a$ by incorporating semantic information provided by the edge map.
First, we compute the difference between the heatmap $H_a$ and the Sobel edge map. This operation can divide $H_a$ with edge information, while it also may disrupt the original $\it{BBOX}$ regions. To ensure the original $\it{BBOX}$ regions are completely preserved in the result, we use the following operation: 
\begin{equation}
\label{He}
H_e(x,y) = \max \big(H_a(x,y) - \lambda Sobel(I)(x,y), H_{\it BBOX}(x,y)\big),
\end{equation}
where $I$ is the text-erased image, and $Sobel$ is the Sobel edge detection operation. $\lambda$ is the weight required to balance the segmentation degree. $H_{\it BBOX}$ is a heatmap in which pixels inside the valid text BBoxes are set to 1.0; otherwise, 0.

Then, the heatmap $H_{e}$ is further segmented using thresholding and we obtain $H_t$:
\begin{equation}
\label{Ht}
H_t(x, y)=\left\{ 
\begin{array}{l}
H_e(x, y), \quad \text{if} \  H_e(x, y) > T\\
0, \hspace{1.4cm} \text{otherwise}
\end{array}
\right.
,
\end{equation}
where $T$ denotes a constant threshold. In our implementation, $T$ and $\lambda$ were set to 0.75 and 5.0, respectively.

\begin{figure*}[!t]
\centering
\includegraphics[width=0.8\linewidth]{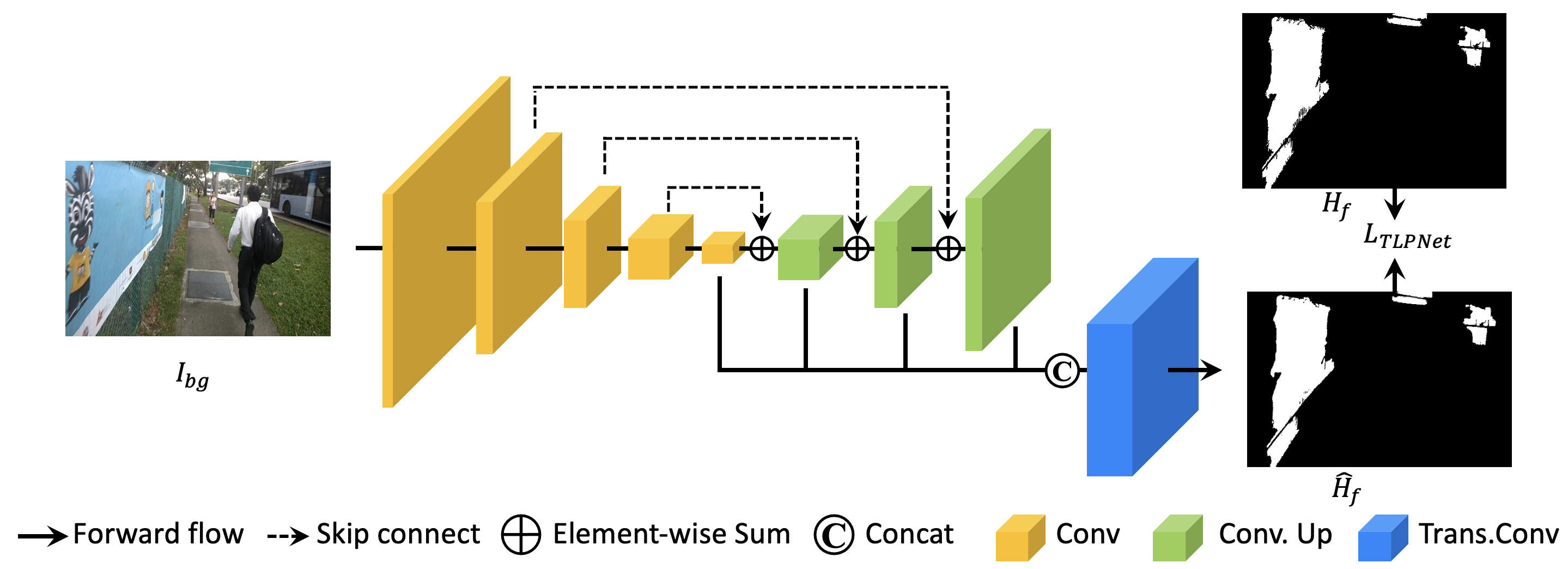}
\caption{Structure of the text location proposal network. Given a background image $I_{bg}$, TLPNet aims to segment the text region, which should be as close to the heatmap $H_f$.}
\label{fig4.1.1}
\end{figure*}
%

Next, we compute all connected components in $H_t$ and mark them as $\mathcal{S}_{j}$, where $j$ is the index of each segmented region. We filter out small regions and regions that do not contain a high appearance consistency score in $\mathcal{S}_{j}$ to ensure final text insert regions are the extension of the $\it{BBOX}$ regions.
Finally, we set the values of pixels inside remaining $\mathcal{S}_{j}$ to 1 and inpaint the small holes to generate the final heatmap $H_f$ as the ground truth for the training of TLPNet. The processing flow of the appearance consistency heatmap is shown in Fig.~\ref{fig4.1.0}. Through our preprocessing, BBox-based text regions are extended into semantic-based ones by considering the similarity of the regions' appearance.

\begin{figure*}[!t]
\centering
\includegraphics[width=0.95\linewidth]{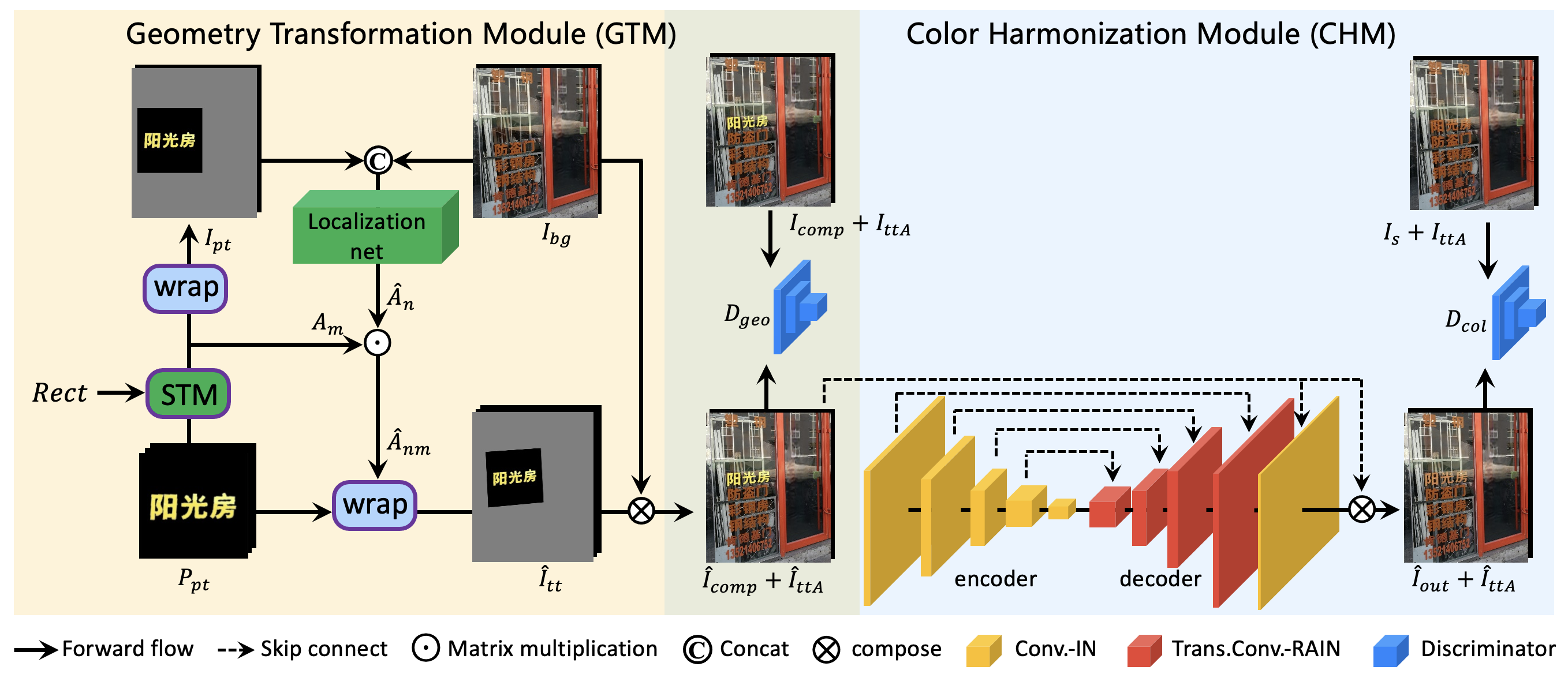}
\caption{Overview of our proposed text appearance adaptation network. It is composed of a geometry transformation module (GTM) (left) and a color harmonization module (CHM) (right). Given an input triplet consisting of a patch plain-text image $P_{pt}$, a reference rectangle $Rect$, and a background image $I_{bg}$, the GTM learns to place the text with a realistic perspective. Then, the CHM takes the composite image $I_{comp}$ and text mask $I_{ttA}$ as inputs, and outputs a synthetic text image $I_{out}$ with a harmonious color.}
\label{fig4.2.0}
\end{figure*}

\subsubsection{Network Structure of TLPNet} 
Given a background image $I_{bg}$, TLPNet aims to segment the mask of the text region $H_{f}$, which is suitable for text embedding. We adopted the segmentation head of the DB \cite{LiaoDB2020} and used ResNeXt-50 \cite{XieResNeXt2017} as the backbone for our TLPNet, which is illustrated in Fig.~\ref{fig4.1.1}. During training, we used a binary cross-entropy (BCE) loss and a DICE loss.

\begin{equation}
\label{BCE loss}
L_{bce}(S,T) = -(T \log(S)+(1-T) \log(1-S) )
\end{equation}
\begin{equation}
\label{Dice} 
L_{dice}(S,T) = 1 - \frac{2 \sum_{i}^N S_{i} T_{i} }{\sum_{i}^N S_{i} + \sum_{i}^N T_{i}}
\end{equation}
\begin{equation}
\label{L_TLPNet}
L_{\it TLPNet} = \lambda_{0} L_{bce}(\hat{H}_{f}, {H}_{f}) + L_{dice} (\hat{H}_{f}, {H}_{f}),
\end{equation}
where $S$ and $T$ represent the prediction and ground truth of the mask image, respectively, and N denotes the total number of pixels in the image. $\hat{H}_{f}$ and ${H}_{f}$ are the prediction and ground truth of TLPNet, respectively. $\lambda_{0}$ is set as 10 in our implementation.

\subsection{Text Appearance Adaptation Network} 
We consider that the realism of text appearance has two aspects: proper perspective and harmonious color that align with the background context. To address this, our TAANet comprises 1) a geometry transformation module (GTM) and 2) a color harmonization module (CHM), as illustrated in Fig. ~\ref{fig4.2.0}. 
For the GTM, there are three inputs: a patch-level plain text image $P_{pt}$, a background image $I_{bg}$, and a reference rectangle $Rect$ indicating the approximate location and scale of the text in the background image. The GTM outputs a composed image $\hat{I}_{comp}$, where $P_{pt}$ is transformed by homography matrices to fit the local geometric context of the background based on $Rect$. $P_{pt}$ is a fixed-size text-centered patch image in which text is placed horizontally. In the CHM, the composed image $\hat{I}_{comp}$ and its corresponding text mask $I_{ttA}$ are taken as inputs, and the output $\hat{I}_{out}$ is an image in which the color of the text is properly transferred to harmonize with the background.

\subsubsection{Data Preprocessing} 
In a simple image-level text synthesis scenario, we are provided with a background image, a plain text patch, and a hint indicating the rough location of the text. Accordingly, given the BBox and stroke-level mask of one text instance from the source scene-text image $I_{s}$ in the DecompST dataset, preprocessing aims to remove the original geometry and color information of text instances to obtain a patch-level plain text image $P_{pt}$ and a reference rectangle $Rect$. $P_{pt}$ is a text instance with a perturbed color and horizontal layout, while $Rect$ indicates the approximate location and scale of the text within $I_{s}$. By restoring $P_{pt}$, $Rect$, and the text-erased image $I_{bg}$ back to $I_{s}$, the geometry and color relationship between text and background can be learned through TAANet.

\begin{figure}[!t]
\includegraphics[width=\linewidth]{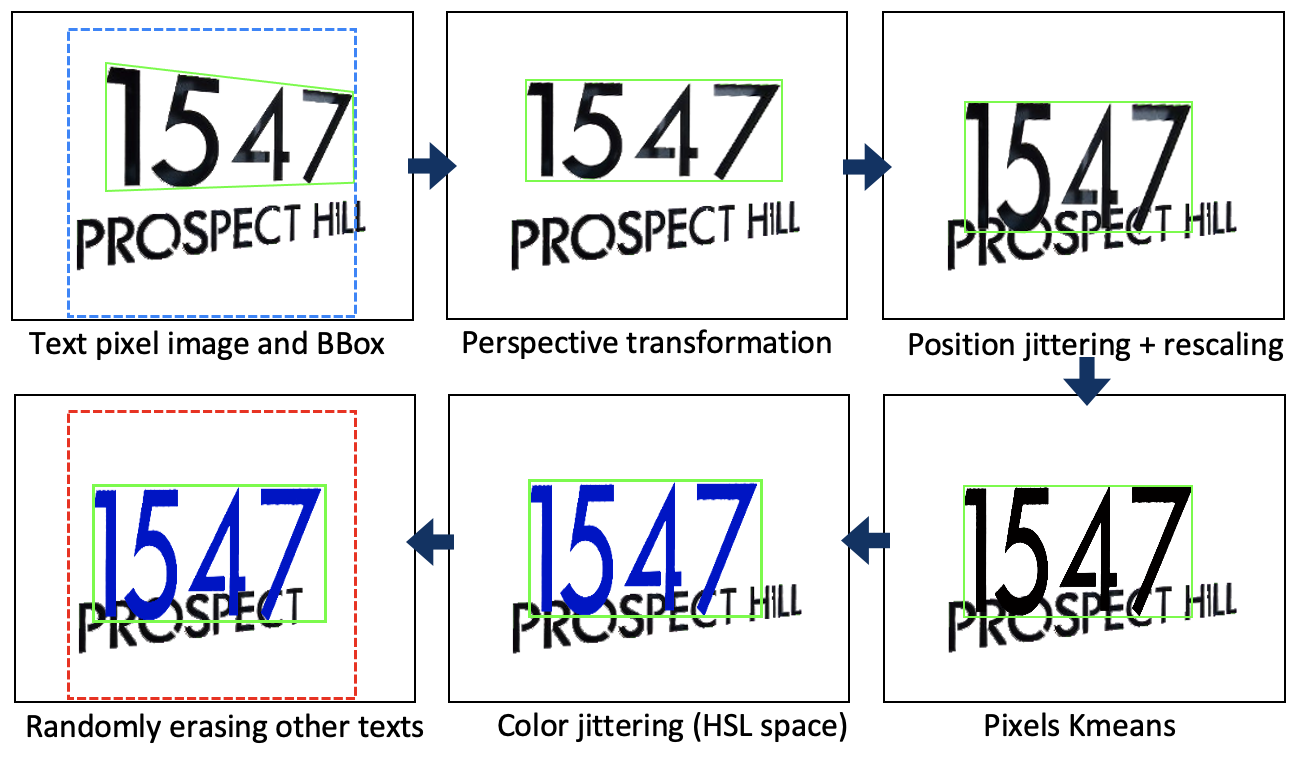}
\caption{Flow of the preprocessing of training data in TAANet. The \textcolor{blue}{blue} and \textcolor{red}{red} dashed boxes are the same reference rectangle $Rect$ but in the images before and after the processing to show the clipping regions to obtain patch text images.}
\label{fig4.2.1}
\end{figure}

The first step of preprocessing is to cut off the target text instance from the text-pixel image and apply a perspective transformation to warp the target text instance into a rectangular one based on its quadrilateral-BBox annotation so that we obtain a horizontal text instance without perspective. Sequentially, we augment the data by randomly altering the aspect ratio of the rectangle BBox and jittering the center of the rectangle BBox, to further perturb the geometric relationship between the target text instance and the background. Next, the text pixels of the target text instance are clustered in only two or three colors using K-means to remove color information and noise. In addition, we augment the data by jittering the color of the text in the HSL space. Finally, to reduce the interdependence between text instances within an image, other text instances are randomly erased in the background image. The entire process flow is shown in Fig.~\ref{fig4.2.1}.

%
\begin{figure}[!t]
\includegraphics[width=\linewidth]{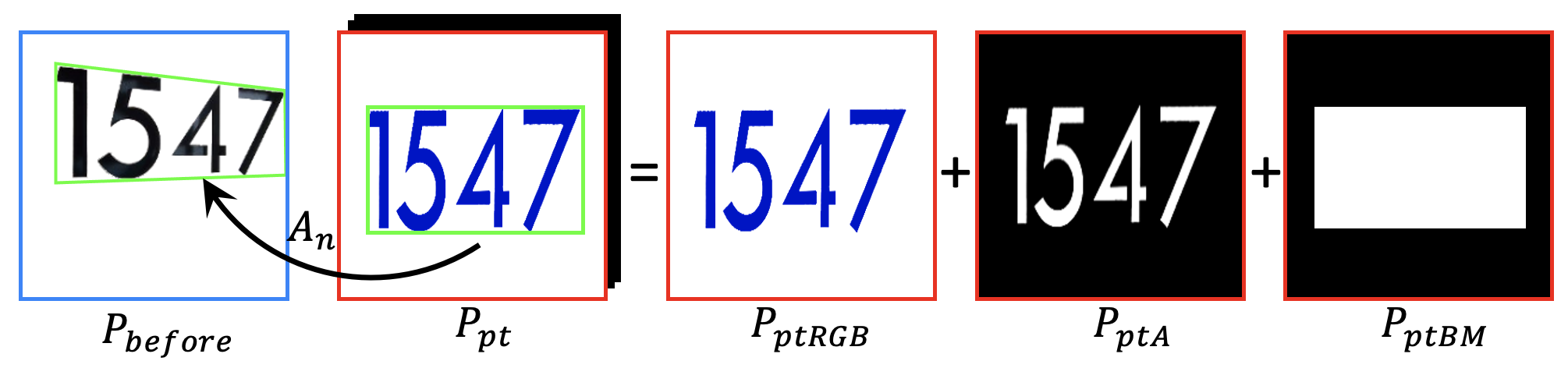}
\caption{Illustration of $A_n$ and $P_{pt}$. The images in \textcolor{blue}{blue} box $P_{\it before}$ and \textcolor{red}{red} box $P_{pt}$ are obtained by cropping from the dash boxes with the same color in Fig.~\ref{fig4.2.1}.}
\label{fig4.2.1-1}
\end{figure}

Based on the aforementioned processing, we can obtain the reference rectangle $Rect$, patch-level plain text image $P_{pt}$, background image $I_{bg}$, and ground truth of the transformation matrix $A_n$ using the following operations. $Rect$ is a square box centered on the processed target text instance. Using $Rect$, the target text instances before and after processing are cropped, resized, and padded to create $P_{pt}$ and the text image before processing $P_{\it before}$. $A_n$ is computed based on the transformed BBox in $P_{pt}$ and the original BBox in $P_{\it before}$. Moreover, $P_{pt}$ is a five-channel image with RGB channels $P_{ptRGB}$, alpha channel $P_{ptA}$, and a mask channel of the BBox-level of the text region $P_{ptBM}$, as shown in Fig.~\ref{fig4.2.1-1}. $P_{ptBM}$ is utilized as additional information during network training, which will be discussed in the later section. Finally, $I_{bg}$ is generated by composing the remaining text in the processed text-pixel image and text-erased image.

\subsubsection{Geometry Transformation Module (GTM)} 
The first step of the GTM is to feed $Rect$ into a spatial transformer module (STM) \cite{ZhanACCV2020} and generate a transformation matrix $A_m$ parameterized by $\theta_m$. The $A_m$ is used to warp and pad the patch-level plain text image $P_{pt}$ into the plain-text image $I_{pt}$. Then, the $I_{pt}$ and background image $I_{bg}$ are concatenated and fed into the localization network (ResNet-34 \cite{HeResNet2016}) to regress the parameters $\theta_n$ of the homography transformation matrix $A_n$. Once the transformation matrices $A_m$ and $A_n$ are obtained, they are applied to the $P_{pt}$ to sample the transformed text image $I_{tt}$. In the GTM, $A_m$ is used to determine the coarse location and scale of the text based on the $Rect$, and $A_n$ is used to transform the local perspective of the text instance. The transformation is expressed as follows:
\begin{equation}
\label{transformation}
\binom{x^{tt}_{i}}{y^{tt}_{i}} = \mathcal{T}_{\theta_m}(\mathcal{T}_{\theta_n}(G_i)) = A_m A_n \left({\begin{array}{c} x^{pt}_{i}\\y^{pt}_{i}\\1\end{array} }\right),
\end{equation}
where $\mathcal{T}_{\theta}$ is a 2D perspective transformation and $G_i$ is a pixel in a regular grid $G$, which is the same as the grid in $P_{pt}$. Therefore, $G_i = (x^{pt}_{i}, y^{pt}_{i})$, which are the coordinates of $P_{pt}$, and $(x^{tt}_{i}, y^{tt}_{i})$ are the corresponding coordinates in the warped grid that defines the sample points.
\begin{equation}
\label{Itt}
I_{tt} = \mathcal{S}(\mathcal{T}_{\theta_m}(\mathcal{T}_{\theta_n}(G)), P_{pt}),
\end{equation}
where $\mathcal{S}$ represents the differentiable bilinear sampler \cite{JaderbergSTN2015} that computes the pixel value of $I_{tt}$ by interpolating the corresponding neighbor pixels in $P_{pt}$. 

After obtaining the transformed text image $I_{tt}$ and background image $I_{bg}$, we can compose them to obtain ${I}_{comp}$:
\begin{equation}
\label{Icomp}
{I}_{\it comp} = I_{\it ttRGB} \circ  I_{ttA} +  I_{bg} \circ (1 - I_{ttA}),
\end{equation}
where ${\circ}$ is the Hadamard product. $I_{\it ttRGB}$ and $I_{ttA}$ are the RGB channels and pixel-level alpha channel of $I_{tt}$.

During the training, we introduce three loss functions to stabilize the training of the geometry transformation module: local L1 loss, global region loss, and adversarial loss. 
We use a robust smooth-L1 loss \cite{girshick_fastRCNN_2015}, as the local L1 loss directly restricts the output of the localization network from a numerical perspective:
\begin{equation}
\label{L_1} 
\begin{aligned}
L_{1} = \text{smooth}_{L1}(\hat A_n - A_n),
\end{aligned}
\end{equation}
\begin{equation}
\label{L_smL1} 
\begin{aligned}
\text{smooth}_{L1}(x) = \left\{ 
\begin{array}{l}
0.5x^2 \ \qquad \text{if} \  |x|<1\\
|x|-0.5 \quad \text{otherwise},
\end{array}
\right.
\end{aligned}
\end{equation}
where $\hat A_n$ and $A_n$ represent the prediction and ground truth of the localization network output, respectively.

The region loss employs the DICE loss in Eq.~\ref{Dice} to guide the transformed text with a higher overlapping rate from the view of the region, and we globally apply it to the stroke-mask level and BBox-mask level in the image:
\begin{equation}
\label{L_Gdice} 
\begin{aligned}
L_{\it region} = L_{dice} (\hat{I}_{ttA}, I_{ttA}) + L_{dice} (\hat{I}_{\it ttBM}, {I}_{\it ttBM}),
\end{aligned}
\end{equation}
here, $\hat{I}_{ttA}$ and $\hat{I}_{\it ttBM}$ are generated by transforming ${P}_{ptA}$ and ${P}_{\it ptBM}$ using the matrices $A_m$ and $\hat A_n$. $I_{ttA}$ and ${I}_{\it ttBM}$ are the corresponding ground truths that can be easily generated from stroke-level text masks and text BBoxes.

GAN \cite{LinSTGAN2018,ZhanSFGAN2019,ChenGCC2019} has been proven beneficial for the training of STN, so we adopt it in our implementation. However, we do not directly use the source image $I_s$ as the ``real image" in adversarial training because $I_s$ is realistic in both the geometry and color spaces. Instead, we generate $I_{\it comp}$ by warping $I_{pt}$ using $A_n$, which only achieves realism in the geometry domain. $I_{\it comp}$ is treated as a ``real image" during the training of the GTM. The adversarial loss is defined as follows:

\begin{equation}
\label{L_Dgeo} 
\begin{aligned}
L_{D_{\it geo}} = &\mathbb{E}_{I_{\it comp}}[\text{ReLU}(1-D_{\it geo}(I_{\it comp}, I_{ttA}))] \\
                        &+ \mathbb{E}_{\hat I_{\it comp}}[\text{ReLU}(1+D_{\it geo}(\hat I_{\it comp},\hat I_{ttA}))]
\end{aligned}
\end{equation}

\begin{equation}
\label{L_G} 
\begin{aligned}
L_{\it GTM} = &\lambda_{1}L_{1} + \lambda_{2}L_{\it region}  \\
                    &- \mathbb{E}_{\hat I_{\it comp}}[D_{geo}(\hat I_{\it comp},\hat I_{ttA})],
\end{aligned}
\end{equation}
where $I_{\it comp}$ and $I_{ttA}$ are concatenated as the inputs of the discriminator. $\lambda_{1}$ and $\lambda_{2}$ are set to 50 and 10, respectively, in our experiment.

\begin{figure*}[!t]
\centering
\begin{minipage}[t]{0.245\linewidth}
    \includegraphics[height=0.75\textwidth, width=1\textwidth]{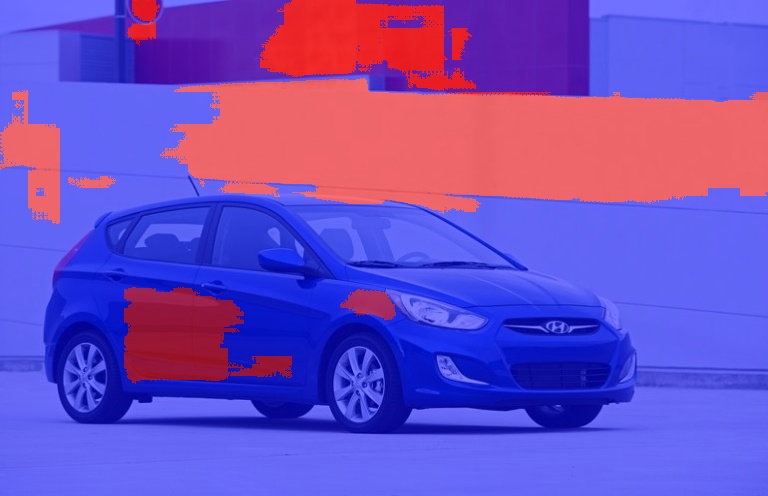}\vspace{2pt}
    \includegraphics[height=0.75\textwidth, width=1\textwidth]{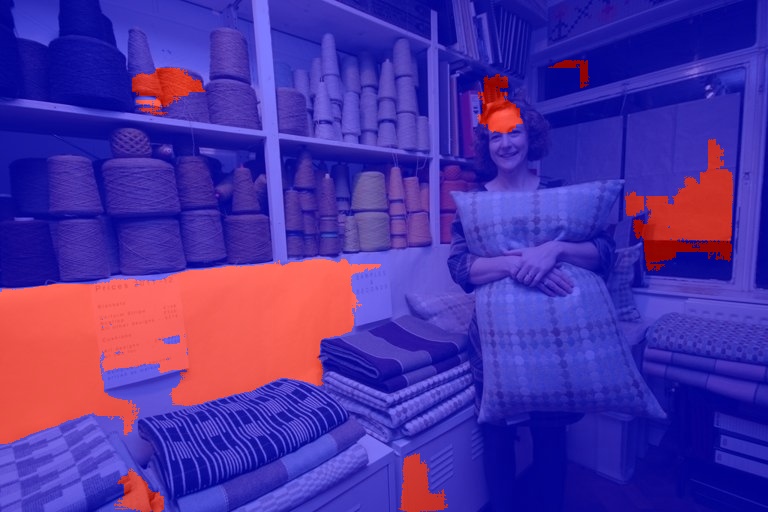}\vspace{2pt}
    \includegraphics[height=1.2\textwidth, width=1\textwidth]{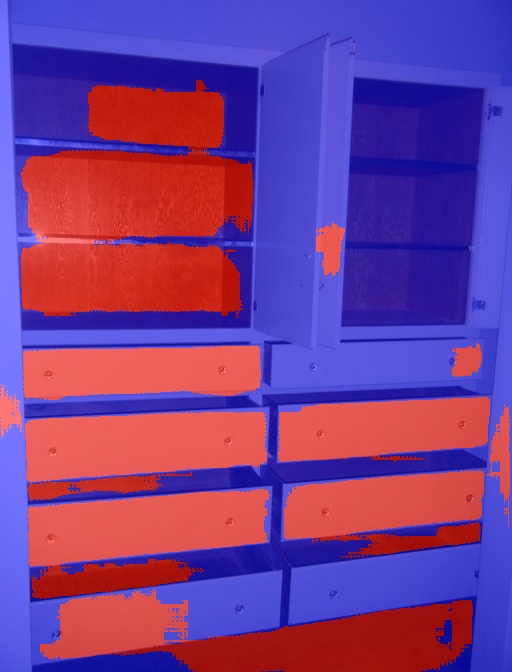}\vspace{4pt}
\end{minipage}
\hspace*{-5pt}
\begin{minipage}[t]{0.245\linewidth}
    \includegraphics[height=0.75\textwidth, width=1\textwidth]{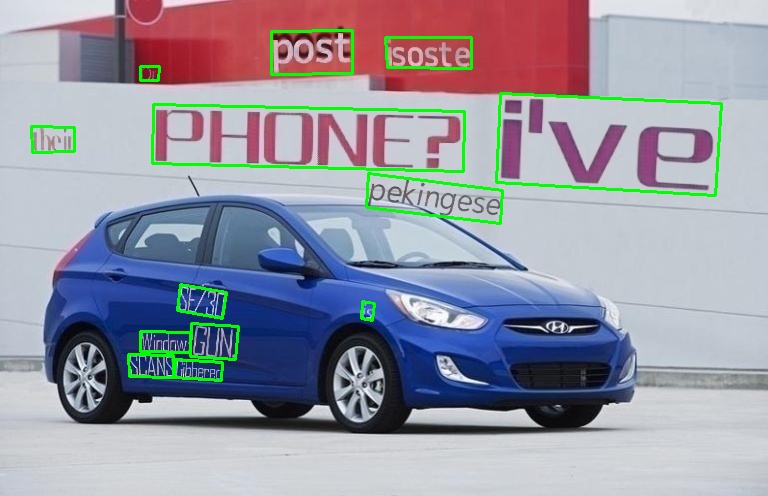}\vspace{2pt}
    \includegraphics[height=0.75\textwidth, width=1\textwidth]{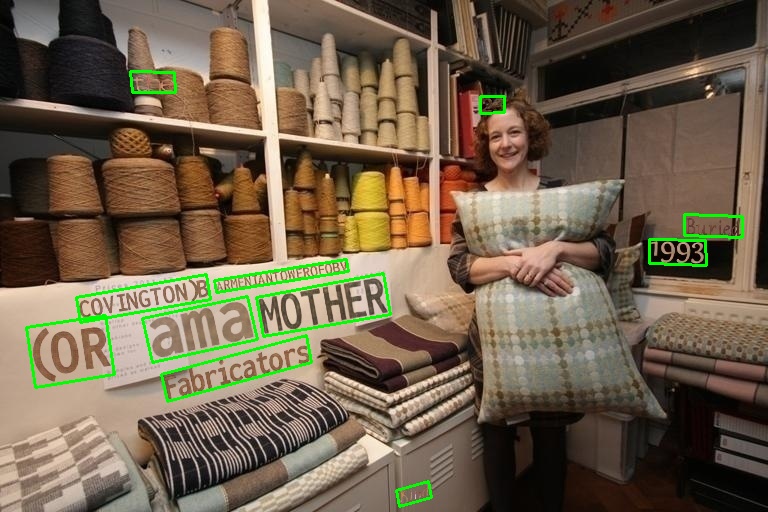}\vspace{2pt}
    \includegraphics[height=1.2\textwidth, width=1\textwidth]{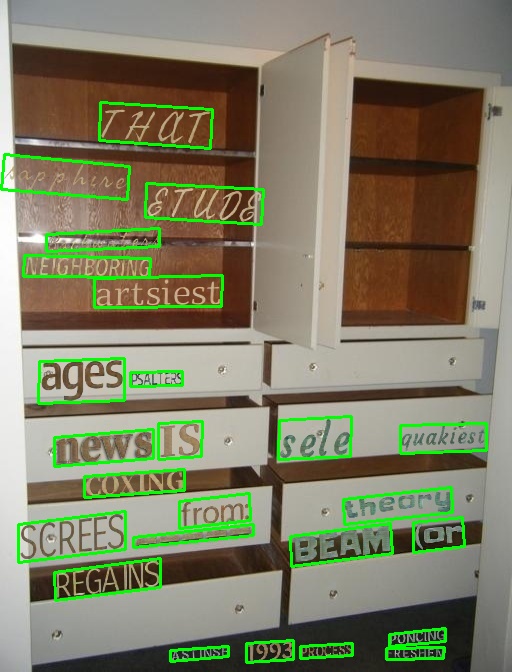}\vspace{4pt}
\end{minipage}
\vspace{1pt}
\begin{minipage}[t]{0.245\linewidth}
    \includegraphics[height=0.75\textwidth, width=1\textwidth]{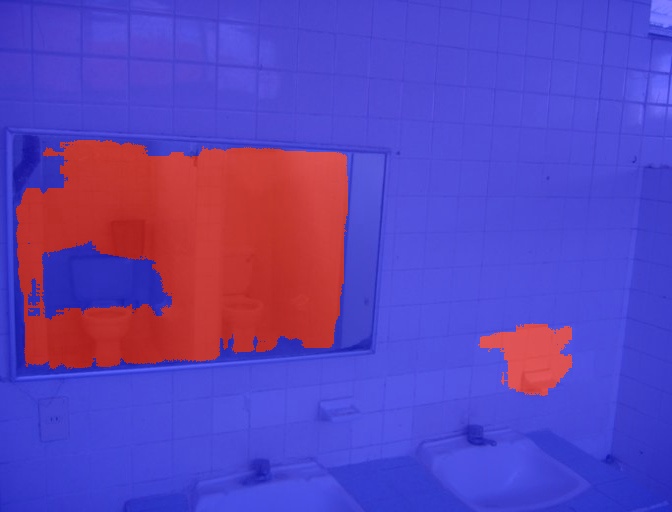}\vspace{2pt}
    \includegraphics[height=0.75\textwidth, width=1\textwidth]{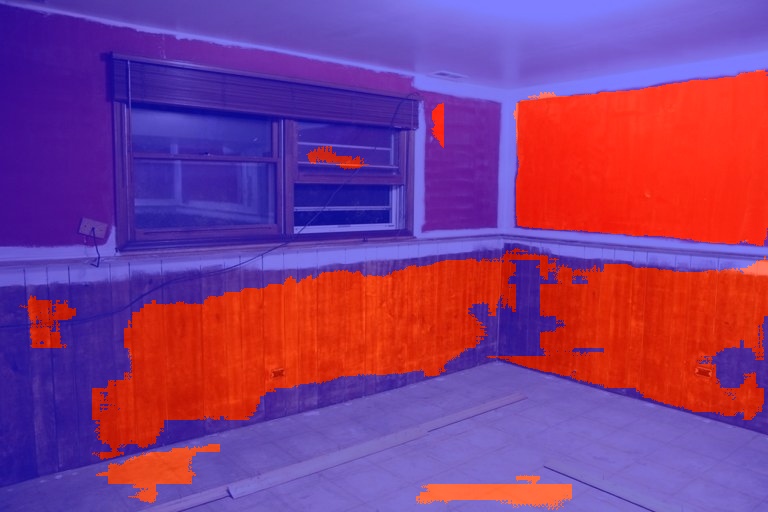}\vspace{2pt}
    \includegraphics[height=1.2\textwidth, width=1\textwidth]{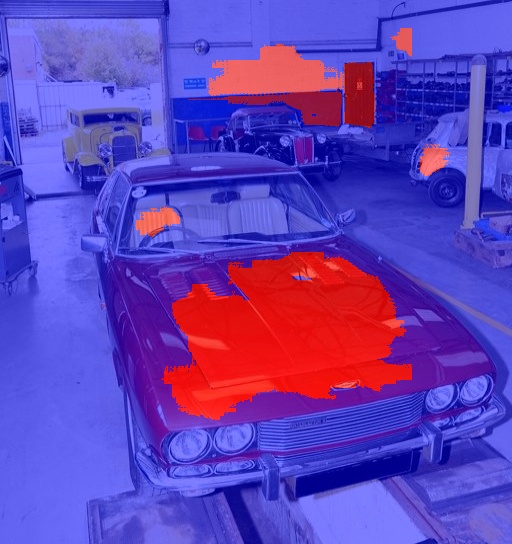}\vspace{4pt}
\end{minipage}
\hspace*{-5pt}
\begin{minipage}[t]{0.245\linewidth}
    \includegraphics[height=0.75\textwidth, width=1\textwidth]{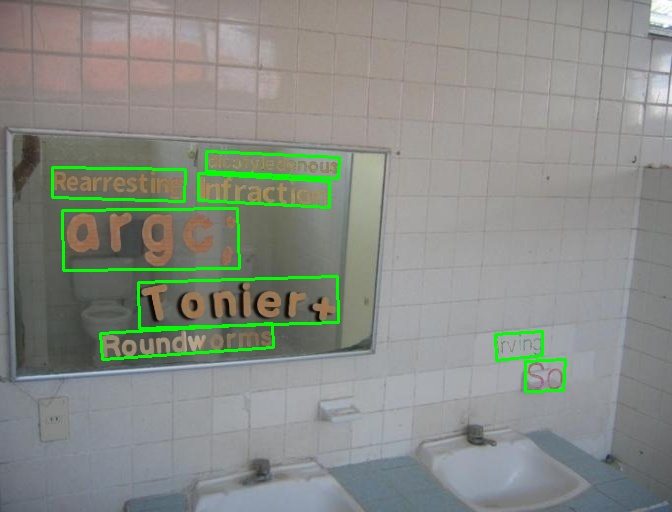}\vspace{2pt}
    \includegraphics[height=0.75\textwidth, width=1\textwidth]{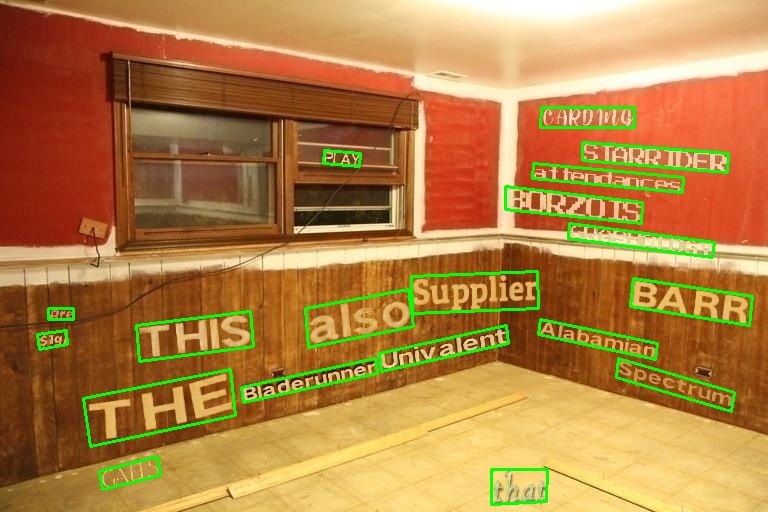}\vspace{2pt}
    \includegraphics[height=1.2\textwidth, width=1\textwidth]{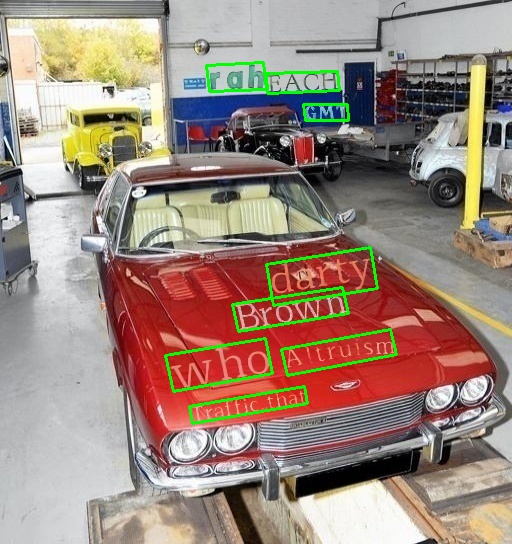}\vspace{4pt}
\end{minipage}
\vspace{-1pt}
\caption{Several sample images generated by our proposed synthesis engine. The left column of the paired images displays the predicted text regions using TLPNet, and the right column of that is our synthesized image.}
\label{fig5.1.2}
\end{figure*}

\subsubsection{Color Harmonization Module (CHM)} 
We treat this text-color-changing task as an image-harmonization problem. We employ the region-aware adaptive instance normalization (RAIN) module \cite{LingRAIN2021} in a UNet-like architecture by adding RAIN modules after the convolutional layers in the decoding stage. 
RAIN is proposed as an activation function that normalizes the foreground features and aligns the normalized features with a computed scale and bias from the background features. In our task, we hope that it can transfer the style from the background into text instances, maintaining harmony between texts and the background. Given an input feature batch $F \in \mathbb{R}^{C\times H\times W}$ and resized foreground (text) mask $M\in \mathbb{R}^{H\times W}$, the formulation of RAIN$(\cdot)$ is expressed as:
\begin{equation}
\label{RAIN} 
\begin{aligned}
\text{RAIN}(F, M) = \sigma(F, 1-M) \Big(\frac{F-\mu(F, M)}{\sigma(F, M)}\Big) + \mu(F, 1-M),
\end{aligned}
\end{equation}
where $\mu(\cdot)$ and $\sigma(\cdot)$ $ \in \mathbb{R}^{C}$ are the channel-wise mean and standard deviation of the foreground or background features, respectively, computed independently across spatial dimensions for each channel.
\begin{equation}
\label{mu} 
\begin{aligned}
\mu_c(F, M)= \frac{1}{\sum\limits_{h,w} M} \sum_{h,w} F_{c,h,w} \circ M_{h,w}
\end{aligned}
\end{equation}
\begin{equation}
\label{sigma} 
\begin{aligned}
\sigma_c(F, M) = \sqrt{\frac{1}{\sum\limits_{h,w} M} \sum_{h,w}( F_{c,h,w} \circ M_{h,w}-\mu_c(F,M))^2+\epsilon},
\end{aligned}
\end{equation}
where $\circ$ denotes the Hadamard product.

In addition, we adopt an adversarial training method. Adversarial loss can be expressed as follows:
\begin{equation}
\label{L_Dcol} 
\begin{aligned}
L_{D_{\it col}} =& \mathbb{E}_{I_{s}}[\text{ReLU}(1-D_{\it col}(I_{s}, I_{ttA}))] \\
&+ \mathbb{E}_{\hat I_{\it out}}[\text{ReLU}(1+D_{\it col}(\hat I_{out},\hat I_{ttA}))]
\end{aligned}
\end{equation}

\begin{equation}
\label{L_Gcol} 
\begin{aligned}
L_{\it CHM} = \lambda_{3} \| \hat I_{out} - I_{s}\| -\mathbb{E}_{\hat I_{out}}[D_{col}(\hat I_{out},\hat I_{ttA})].
\end{aligned}
\end{equation}
Here, $\lambda_{3}$ is set to 5 in the experiment.

\subsubsection{Inference Pipeline}
After training the TLPNet and TAANet, they can be integrated into a generation pipeline to generate synthetic data. The inference process of our method is illustrated in the lower section of Fig.~\ref{fig4.0}. Given a background image, We first use TLPNet to predict the text regions in the form of heatmaps. Subsequently, we randomly sample a reference rectangle with a higher 70$\%$ overlap rate with the text regions. At the same time, a plain text patch image with a size of 256 $\times$ 256 is generated by randomly selecting fonts, text, and color. Then, the reference rectangle, plain text patch image, and background image are passed through TAANet to produce a synthetic text image. Finally, post-processing applies various effects to the text, including shadows, 3D effects, texture, and blurring. In the composition of the multiple text instances within one background image, we abandon overlapped and small text instances. In the presence of semantic information, such as in the COCO dataset \cite{MSCOCO2014}, the refinement of the synthesis can be achieved by discarding the text beyond the boundaries of semantic segmentation, allowing for the synthesis of text instances specifically on designated objects.

\section{Experiment} 
\label{Experiment}
\subsection{Implementation Details} 
\subsubsection{Training Configurations}
Our implementation was based on the PyTorch framework. For training of TLPNet, we used the DecompST and the SCUT-EnsText datasets \cite{LiuEraseNet2020} to generate the training data pairs. As a result, we obtained a total of approximately 7900 training data pairs. The input size of the TLPNet was set to 768 $\times$ 768, and the batch size was 12 on an Nvidia GeForce RTX 3090 GPU. We employed the Adam \cite{KingmaAdam2015} optimizer with a $\beta$ of (0.5, 0.9), and the learning rate started at 0.0002 and decayed to nine-tenths after every 20 epochs in the training phase.
During the training of TAANet, GTM and CHM were trained separately. This is because we adopted the L1 loss during the training of CHM, which is essential for effectively constraining the color of the output. The input size of TAANet was also 768 $\times$ 768, and the training batch size for GTM and CHM were set to 20 and 10, respectively, on a single Nvidia GeForce RTX 3090 GPU. The optimizer used was the same as in TLPNet, and the discriminators' learning rate started from 0.0004, with the same decay rate as that in TLPNet.

\subsubsection{Inference Configurations}
In the preparation stage, we need to collect some ingredients for synthesis, including background images, fonts, and a lexicon. The background images were collected from the COCO dataset \cite{MSCOCO2014} and Places2 dataset \cite{zhou_places_2018}. To ensure that the images closely resembled real scene images, we selected the image sets by excluding those with labels related to natural landscapes. Additionally, we applied filtering to the selected image sets using CRAFT \cite{BaekCRAFT2019} and DB \cite{LiaoDB2020} to remove any images with prominent text. Ultimately, we amassed a collection of approximately 200,000 background images. Furthermore, we gathered around 2000 fonts and compiled a lexicon by combining the MJ dataset \cite{JaderbergMJ2016} and the ST dataset \cite{gupta_synthetic_2016}. 
Our LBTS dataset is generated by a machine with a single GeForce RTX 3080 GPU, AMD Ryzen7 3700X @ 3.6 GHz CPU, and 32G RAM. The TLPNet model consists of 24.7M parameters, while the TAANet model has 38.5M parameters (21.4M for GTM and 17.1M for CHM). The inference times for TLPNet on a single image and TAANet on one text instance are approximately 21ms and 81ms (11ms for GTM, and 70ms for CHM), respectively. Fig.~\ref{fig5.1.2} shows some generated samples from our LBTS dataset. We observed that the TLPNet exhibited a preference for predicting the text region in relatively flat areas, especially in regions with quadrilateral shapes. This tendency may stem from the bias in the training data, where most text instances exist on the signs, walls, or billboards. On the other hand, the geometry and color relationship between text and background is also reasonably aligned by the TAANet. The text perspective accurately follows the boundaries of text regions, and the text color is appropriately balanced, neither being obtrusive nor excessively dull.

\begin{table*}[!t]
\centering
\caption{Comparison between Previous Synthetic Datasets and Our LBTS Dataset on the ICDAR2013, ICDAR2015, ICDAR2017MLT Datasets Using EAST as the Baseline Detector. R: recall, P: precision, F: F-score, Real: the corresponding training set of the evaluation dataset}
\label{tab5.3}
\begin{tabular}{cccccccccc} 
\toprule
\multirow{2}{*}{Train dataset} & \multicolumn{3}{c}{IC13}                         & \multicolumn{3}{c}{IC15}                         & \multicolumn{3}{c}{MLT17}                         \\ 
\cmidrule(lr){2-4}\cmidrule(lr){5-7}\cmidrule(lr){8-10}
                               & R              & P              & F              & R              & P              & F              & R              & P              & F               \\ 
\midrule
ST-10k                         & 71.69          & 73.09          & 72.38          & 50.22          & 64.98          & 56.65          & 40.78          & 55.93          & 47.17           \\
VISD-10k                       & \textbf{75.71} & 74.68          & \textbf{75.19} & \textbf{59.94} & 72.17          & \textbf{65.49} & \textbf{42.58} & 61.19          & \textbf{50.21}  \\
UT-10k                         & 64.2           & \textbf{86.58} & 73.73          & 51.52          & \textbf{77.2}  & 61.8           & 37.36          & \textbf{65.65} & 47.62           \\
LBTS-10k                       & 59.18          & 82.23          & 68.83          & 42.95          & 68.14          & 52.69          & 30.28          & 61.56          & 40.59           \\ 
\midrule
Real                           & 68.22          & 86.66          & 76.34          & 74.58          & 84.32          & 79.15          & 56.09          & 72.94          & 63.41           \\
ST-10k + Real                  & 75.16          & 86.54          & 80.45          & 79.15          & 84.57          & 81.77          & 57.41          & 73.47          & 64.46           \\
VISD-10k + Real                & 75.16          & 89.65          & 81.77          & 79.97          & 85.84          & 82.8           & 56.74          & \textbf{75.06} & 64.62           \\
UT-10k + Real                  & 75.25          & 87.75          & 81.02          & 80.07          & 85.68          & 82.78          & 57.11          & 74.11          & 64.51           \\
LBTS-10k + Real                & \textbf{75.53} & \textbf{89.89} & \textbf{82.08} & \textbf{81.03} & \textbf{86.66} & \textbf{83.75} & \textbf{57.63} & 74.49          & \textbf{64.98}  \\
\bottomrule
\end{tabular}
\end{table*}

\subsection{Evaluation Metrics and Datasets}

\subsubsection{Evaluation metrics}
To verify the effectiveness of different text synthesis methods, a common method is to train the same text detector on different synthesized datasets and evaluate the trained detectors on several test sets of real datasets. The better performance of the text detector indicates a higher quality of the training data, implying a better text synthesis strategy. 
Following previous works \cite{ZhanVISD2018,LongUnreal2020}, synthetic datasets are evaluated from two perspectives: 1) as  \emph{independent training data} for detection models to assess the possibility that whether synthetic datasets can be a substitute for real-world datasets. 2) as \emph{pretraining data} to initialize text detectors, where pretrained models fine-tuned with real-world data usually exhibit better performance than models directly trained from scratch with real-world data.

In our experiment, we selected EAST \cite{zhou_east_2017} and DB \cite{LiaoDB2020} as the baseline text detector to conduct comparison experiments. Both of them were previous state-of-the-art methods and are the most commonly used algorithms in the text detection task. In the implementation of EAST, ResNet-50 \cite{HeResNet2016} was used as the backbone, and all the models were trained on two RTX 2080Ti GPUs with a batch size of 28. For DB, we trained DB-ResNet-50 \cite{LiaoDB2020} on one RTX 3090 with a batch size of 20. The performance metrics of the text detector, recall (R), precision (P), and F-score (F), were calculated under the ICDAR2015 evaluation protocol \cite{KaratzasICDAR2015} over all evaluation datasets.

\subsubsection{Synthetic Dataset} 
\begin{itemize}

\item[\textbullet ] \textbf{Oxford SynthText Dataset (ST)} \cite{gupta_synthetic_2016} is a large-scale synthetic text dataset that consists of about 850,000 images. It is created from about 8000 background images and 1200 fonts. 10,000 data pairs were randomly sampled from this dataset to compose ST-10k.

\item[\textbullet ] \textbf{Verisimilar Image Synthesis Dataset (VISD)} \cite{ZhanVISD2018} contains 10,000 images synthesized from background images collected from the COCO dataset \cite{MSCOCO2014}.

\item[\textbullet ] \textbf{UnrealText (UT)} \cite{LongUnreal2020} initially consists of about 728,000 images in English/Latin. However, we discovered that some of these images either do not contain text or are partially black, potentially due to render failure or incorrect camera positioning. To ensure data quality, we filtered out the images without annotations and those where more than two-thirds of the pixels are completely black. As a result, approximately 670,000 images remained, and we also randomly sampled 10,000 images to form UT-10k for our experiment.
\end{itemize}

\subsubsection{Real-world Dataset}
\begin{itemize}

\item[\textbullet] \textbf{ICDAR 2013 (IC13)} \cite{KaratzasICDAR2013} is a widely used scene text image dataset that includes 229 training images and 233 testing images. 


\item[\textbullet] \textbf{ICDAR 2015 (IC15)} \cite{KaratzasICDAR2015} comprises 1000 training images and 500 test images and addresses incidental scene text in the Latin alphabet. 


\item[\textbullet] \textbf{ICDAR 2017 MLT (MLT17)} \cite{NayefICDAR2017MLT} contains 7200 images for training and 1800 images for validation. Text instances of this dataset are from nine different languages: Arabic, Bangla, Chinese, English, French, German, Italian, Japanese and Korean.


\item[\textbullet] \textbf{Total-Text} \cite{ChngTotalText2017} is a comprehensive dataset of arbitrary-shaped text instances, including horizontal, multi-oriented, and curved textual variations. It contains 1255 training images and 300 test images. All images are annotated with polygons at the word level.

\end{itemize}

\begin{table}[!t]
\centering
\caption{Comparison between Previous Synthetic Datasets and Our LBTS Dataset on the ICDAR2015 and Total-Text Datasets Using DB as the Baseline Detector. R: recall, P: precision, F: F-score, Real: the corresponding training set of the evaluation dataset}
\label{tab5.3-3}
\begin{tabular}{@{}ccccccc@{}}
\toprule
\multirow{2}{*}{Train dataset} & \multicolumn{3}{c}{IC15} & \multicolumn{3}{c}{Total-Text} \\ 
\cmidrule(lr){2-4}\cmidrule(lr){5-7}  
                               & R      & P      & F      & R        & P        & F        \\ \midrule
ST-10k                         & 46.89          & 70.84          & 56.43          & 39.64          & 69.3           & 50.43          \\
VISD-10k                       & \textbf{57.68} & 71.57          & \textbf{63.88} & \textbf{43.66} & \textbf{73.31} & \textbf{54.73} \\
UT-10k                         & 54.94          & \textbf{72.26} & 62.42          & 41.4           & 63.5           & 50.12          \\
LBTS-10k                       & 37.07          & 65.98          & 47.47          & 41.17          & 66.23          & 50.78          \\ \midrule
Real                           & 82.23          & 86.88          & 84.49          & 82.39          & 85.12          & 83.73          \\
ST-10k + Real                  & 82.67          & 89.99          & 86.17          & \textbf{83.7}  & 87.25          & 85.44          \\
VISD-10k + Real                & 83.1           & 89.29          & 86.08          & 82.71          & 87.74          & 85.15          \\
UT-10k + Real                  & 82.91          & \textbf{90.11} & 86.36          & 83.3           & 87.86          & 85.52          \\
LBTS-10k + Real                & \textbf{84.59} & 89.32          & \textbf{86.89} & 82.12          & \textbf{89.3}  & \textbf{85.56} \\ \bottomrule
\end{tabular}
\end{table}

\subsection{Comparison with State-of-the-Art Methods} 
To verify the effectiveness of the proposed text synthesis engine, we conducted evaluation experiments to compare our generated LBTS dataset with those of recent state-of-the-art approaches \cite{gupta_synthetic_2016,ZhanVISD2018,LongUnreal2020}.
First, we standardized the total number of each synthesis dataset to 10k to conduct a fair comparison experiment. We trained EAST on each synthetic dataset with 200,000 steps, followed by fine-tuning on the corresponding real-world training set for an additional 200,000 steps. The performance of EAST was evaluated by the validation set of each real dataset every 1000 steps, and the best F-scores are recorded in Table~\ref{tab5.3}. 
For all the evaluation benchmarks, when we employed synthetic datasets as independent training data, EAST trained on VISD-10k achieved the highest F-score and Recall, and EAST trained on UT-10k achieved higher Precision. However, when we fine-tuned the pretrained EAST with real-world data, we observed that our LBTS-10k dataset outperformed all other synthetic datasets, obtaining 0.31$\%$, 0.95$\%$, and 0.36$\%$ improvement of the F-score on IC13, IC15, and MLT17 datasets over VISD-10k. 

\begin{figure*}[!t]
\centering
\subfigure[EAST (from scratch)]{
\begin{minipage}[t]{0.245\linewidth}
    \includegraphics[height=0.75\textwidth, width=1\textwidth, trim=350 300 0 0, clip]{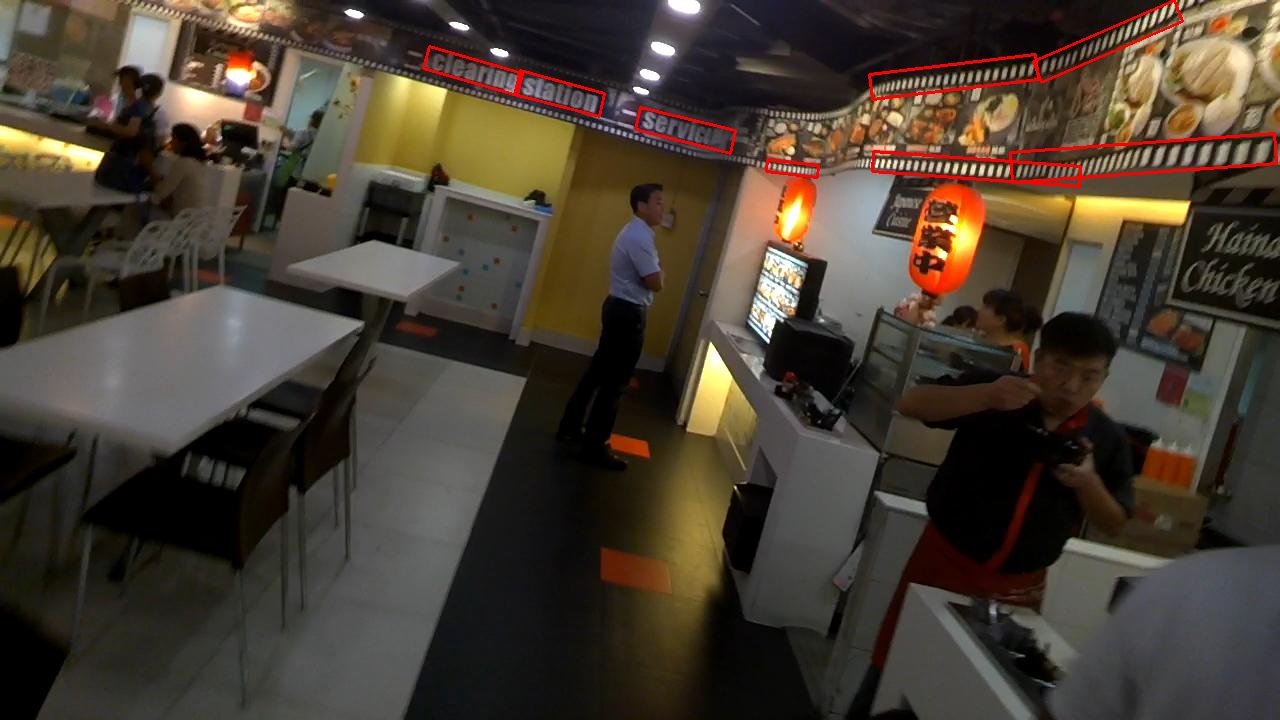}\vspace{2pt}
    \includegraphics[height=0.75\textwidth, width=1\textwidth, trim=0 200 250 100, clip]{fig5.3.1/img_312_1.jpg}\vspace{4pt}
\end{minipage}}
\hspace*{-8.5pt}
\subfigure[EAST (Ours)]{
\begin{minipage}[t]{0.245\linewidth}
    \includegraphics[height=0.75\textwidth, width=1\textwidth, trim=350 300 0 0, clip]{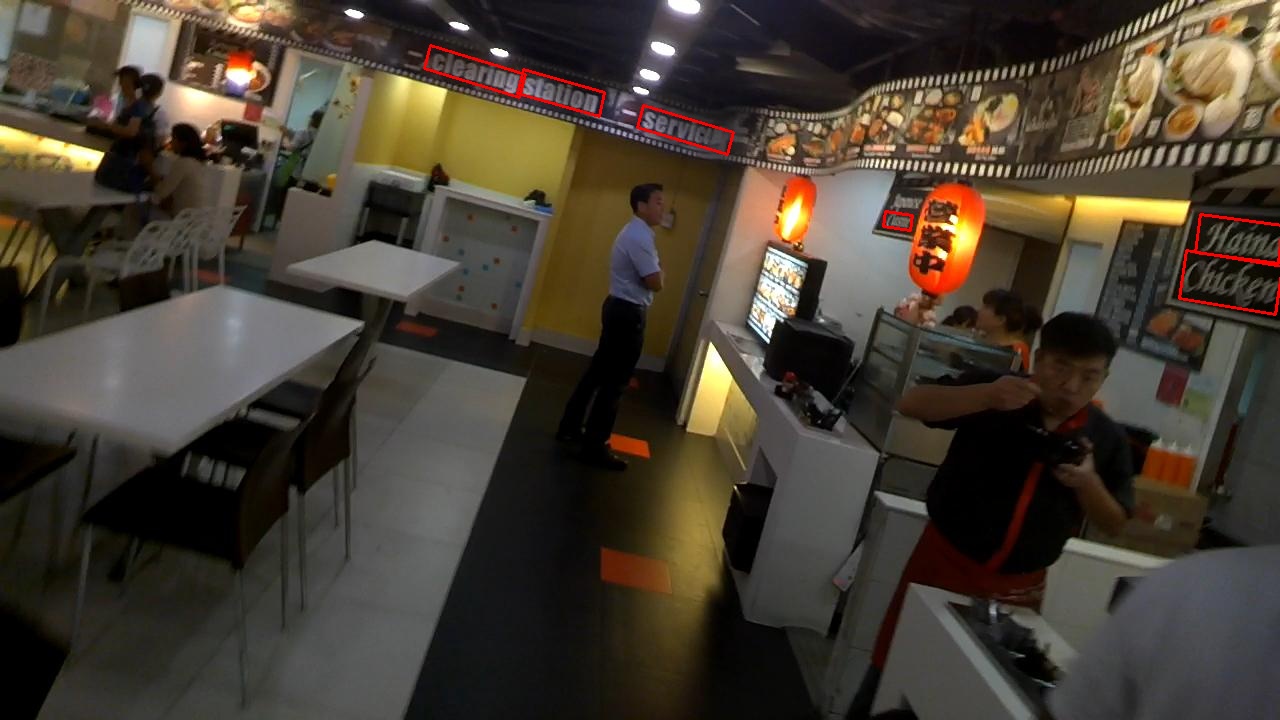}\vspace{2pt}
    \includegraphics[height=0.75\textwidth, width=1\textwidth, trim=0 200 250 100, clip]{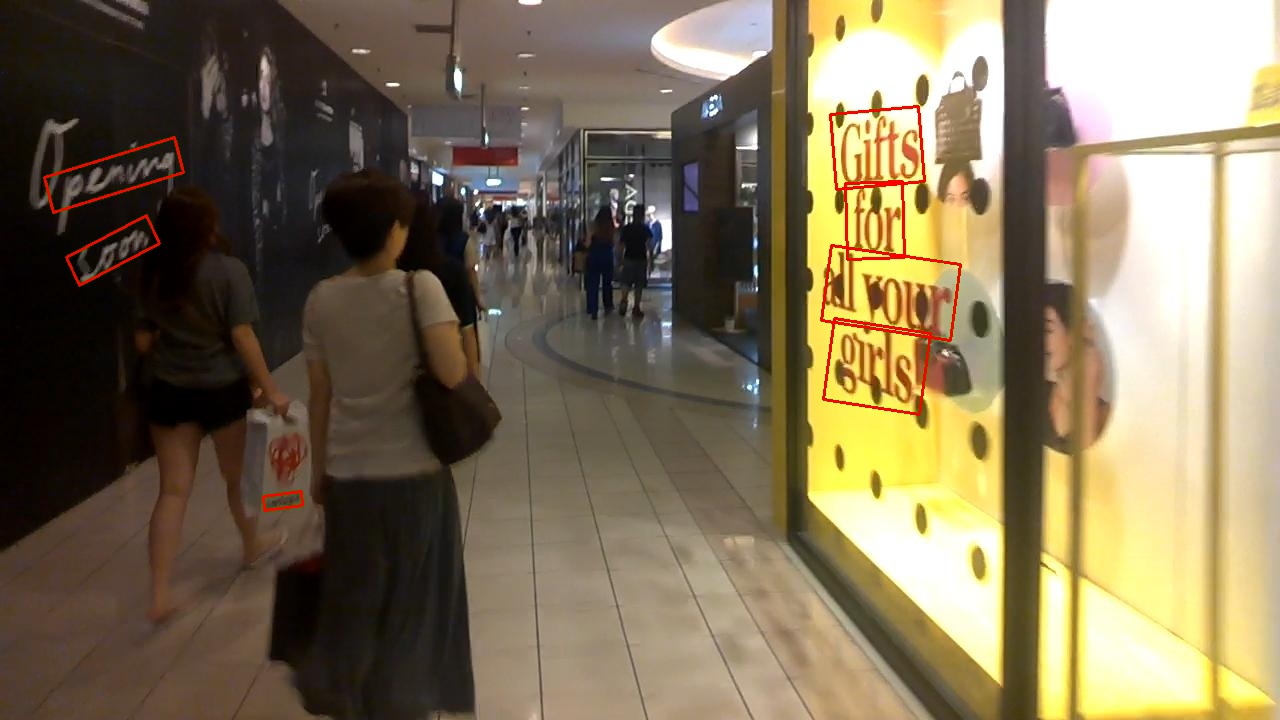}\vspace{4pt}
\end{minipage}}
\hspace*{-7pt}
\subfigure[DB (from scratch)]{
\begin{minipage}[t]{0.245\linewidth}
    \includegraphics[height=0.75\textwidth, width=1\textwidth, trim=0 250 0 100, clip ]{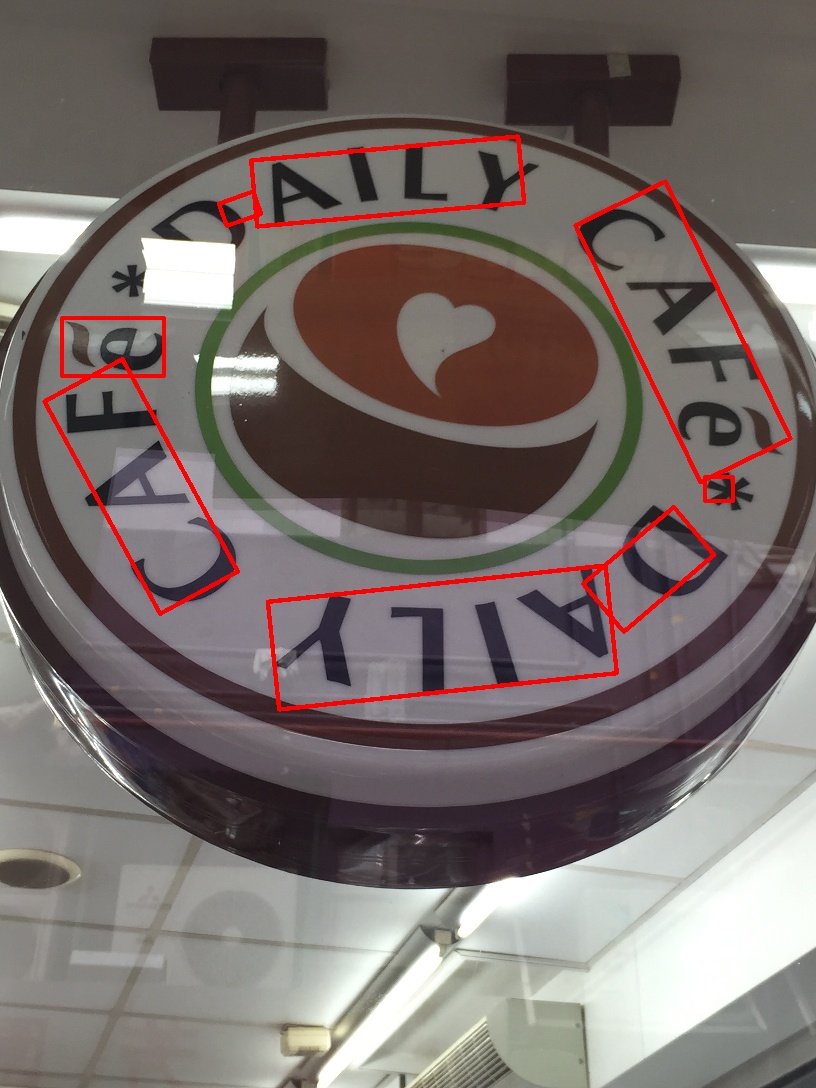}\vspace{2pt}
    \includegraphics[height=0.75\textwidth, width=1\textwidth]{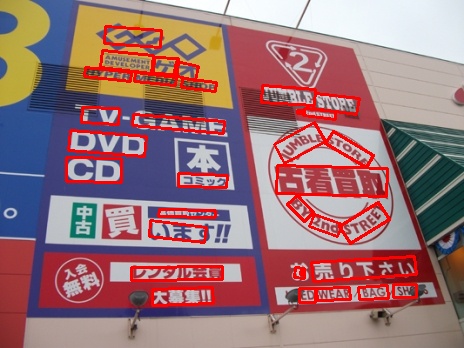}\vspace{4pt}
\end{minipage}}
\hspace*{-8.5pt}
\subfigure[DB (Ours)]{
\begin{minipage}[t]{0.245\linewidth}
    \includegraphics[height=0.75\textwidth, width=1\textwidth, trim=0 250 0 100, clip ]{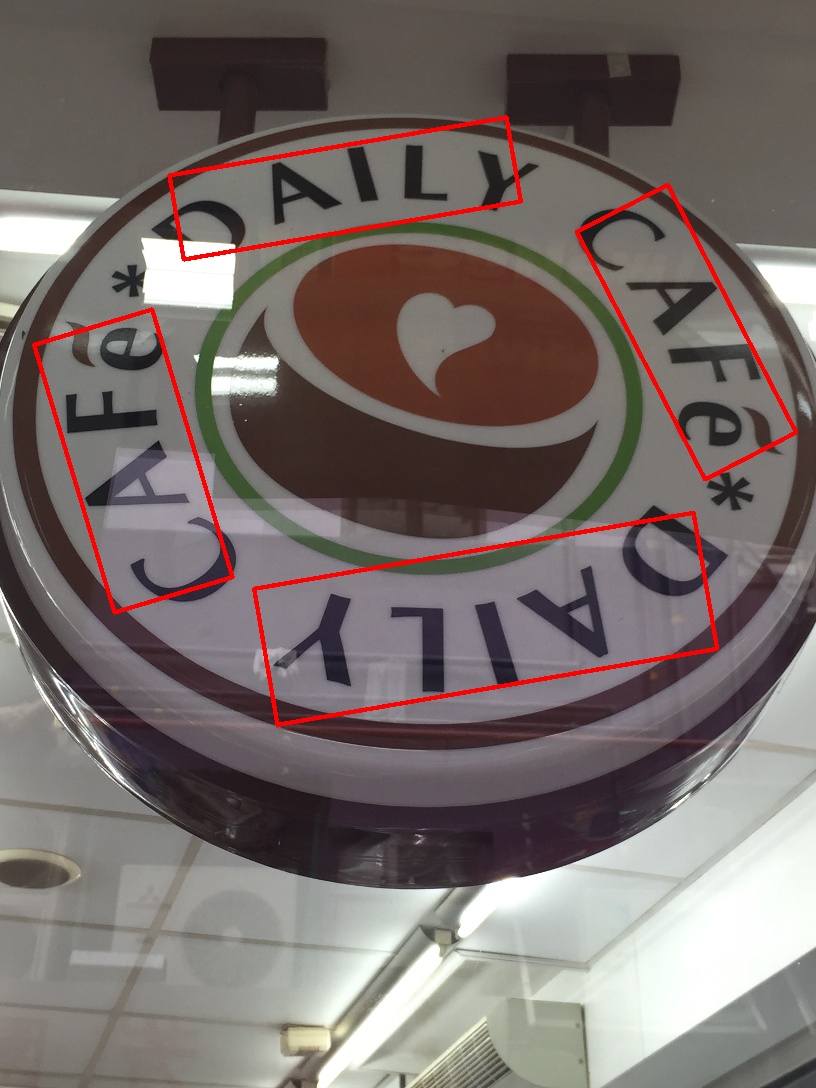}\vspace{2pt}
    \includegraphics[height=0.75\textwidth, width=1\textwidth]{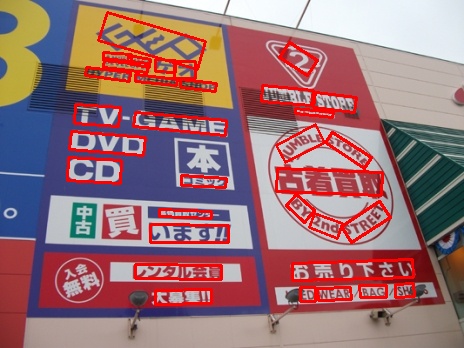}\vspace{4pt}
\end{minipage}}
\vspace{-1pt}
\caption{Visual comparisons of baseline detectors using pretraining. (a) Detection results of EAST trained from scratch. (b) Detection results of EAST pretrained with our LBTS dataset. (c) Detection results of DB trained from scratch. (d) Detection results of DB pretrained with our LBTS dataset. Zoom in for the best view.}
\label{fig5.3.1}
\end{figure*}

\begin{table*}[!t]
\centering
\caption{Quality Comparison between Different Mixed Synthetic Datasets on ICDAR2013, ICDAR2015, ICDAR2017MLT Datasets Using EAST as the Baseline Detector. R: recall, P: precision, F: F-score, Real: the corresponding training set of the evaluation dataset}
\label{tab5.3-1}
\begin{tabular}{cccccccccc}
\toprule
\multirow{2}{*}{Train dataset}           & \multicolumn{3}{c}{IC13}                        & \multicolumn{3}{c}{IC15}                        & \multicolumn{3}{c}{MLT17}                        \\ 
\cmidrule(lr){2-4}\cmidrule(lr){5-7}\cmidrule(lr){8-10}
                                         & R              & P              & F              & R              & P              & F              & R              & P              & F              \\ 
                                         \midrule
VISD-5K + UT-5k                          & 74.7           & 81.39          & \textbf{77.9}  & \textbf{65.86} & 74.59          & \textbf{69.96} & 43.94          & 63.29          & 51.87          \\
VISD-5k + LBTS-5k                        & 74.16          & 79.07          & 76.53          & 60.66          & 72.16          & 65.92          & 42.51          & 62.53          & 50.61          \\
UT-5k + LBTS-5k                          & 69.41          & \textbf{84.35} & 76.15          & 58.02          & \textbf{75.22} & 65.51          & 40.65          & \textbf{65.32} & 50.11          \\
VISD-3.3K + UT-3.3k + LBTS-3.3k          & \textbf{75.98} & 77.9           & 76.93          & 64.42          & 74.71          & 69.18          & \textbf{45.22} & 61.25          & \textbf{52.03} \\ 
\midrule
VISD-5k + UT-5k + Real                   & 75.16          & 88.69          & 81.36          & 78.86          & 86.48          & 82.5           & 57.6           & 73.84          & 64.71          \\
VISD-5k + LBTS-5k + Real                 & \textbf{76.44} & 90             & \textbf{82.67} & \textbf{80.65} & \textbf{87.01} & \textbf{83.71} & 57.55          & \textbf{74.47} & \textbf{64.93} \\
UT-5k + LBTS-5k + Real                   & 75.43          & \textbf{90.27} & 82.19          & 80.07          & 86.98          & 83.38          & \textbf{57.57} & 73.94          & 64.74          \\
VISD-3.3k + UT-3.3k + LBTS-3.3k + Real   & 75.53          & 88.35          & 81.44          & 80.16          & 86.63          & 83.27          & 56.45          & 74.4           & 64.19          \\ 
\bottomrule
\end{tabular}
\end{table*}

We also trained DB in a similar manner to compare the quality of synthesis datasets. Initially, DB was pretrained on each synthetic dataset for 100,000 steps and then fine-tuned on the IC15 or Total-text datasets for another 1200 epochs. During the training, we validated the model with the corresponding test set every 2000 steps, and Table~\ref{tab5.3-3} presents the best F-scores obtained. The results showed that using DB as the baseline detector yielded similar results as using EAST. When considering synthetic datasets as independent training data, the VISD-10k achieved the highest F-score for both IC15 and Total-Text datasets. However, by further fine-tuning the DB model, pretrained on synthetic data, with real data, our LBTS-10k dataset obtained a higher F-score than other datasets. 
Compared to the F-score of DB trained from scratch, DB pretrained with our dataset gained 2.4$\%$ and 1.83$\%$ on IC15 and Total-Text, respectively. Furthermore, in comparison to previous state-of-the-art datasets, we observed a commendable improvement of 0.53$\%$ in F-score on IC15, while achieving competitive performance on the Total-Text dataset. 
To verify the robustness of each synthetic dataset, three random samples of 10k data were extracted from each full-size dataset. These sampled 10k datasets were then used to conduct the evaluation experiments on IC15 using DB. The average F-measure for ST-10k, VISD-10k, UT-10k, and LBTS-10k were 86.24, 86.11, 86.39, and 86.78, respectively. The corresponding variances in F-measure were 0.017, 0.004, 0.019, and 0.015, indicating our LBTS datasets achieve consistently high performance across multiple samples.
Fig.~\ref{fig5.3.1} displays some visual comparisons of baseline detectors with and without LBTS pretraining. Pretrained models effectively reduce detection errors and exhibit enhanced robustness in handling complex text instances. 


To the best of our knowledge, this is the first report that highlights the performance discrepancy resulting from the use of synthetic datasets during the pretraining and fine-tuning stages. In our perspective, synthetic datasets play different roles when employed as independent training data or as pretraining data. When text detectors are solely trained on synthetic datasets and evaluated on real datasets, the performance of the text detector indicates the level of entangled ``realism'' between the synthetic dataset and real data to a certain extent. We believe that the realism of text encompasses multiple dimensions, such as text appearance, distribution, font, lighting conditions, and background image types. Both existing methods and our proposed LBTS approach impose constraints on the generated synthesis data in these dimensions to approximate the real-world domain. Those constraints are usually divided into several rules and steps based on prior knowledge. The "realism" we mentioned here denotes the degree of entangled "realism" achieved based on these constraints.

However, when synthetic datasets served as pretraining data, we hypothesize that dataset diversity becomes more crucial than "realism". \cite{kataokapretrain2022} is one extreme case that the models can be well pretrained without natural images. Synthetic data with greater diversity may enable convolutional layers to learn distinctive representations. These representations' corresponding model weights are activated and reinforced if they are beneficial during the fine-tuning phase, thereby preventing the model from becoming trapped in local minima during gradient descent. We consider that the learning mechanism implemented in our LBTS engine introduces a greater degree of diversity compared to rule-based methods, resulting in our generated data performing better as pretraining data.

\begin{table*}[!t]
\centering
\caption{Quality Comparison between Different Full-size Synthetic Datasets on ICDAR2013, ICDAR2015, ICDAR2017MLT Datasets Using EAST as the Baseline Detector. R: recall, P: precision, F: F-score, Real: the corresponding training set of the evaluation dataset}
\label{tab5.3-2}
\begin{tabular}{cccccccccc} 
\toprule
\multirow{2}{*}{Train dataset} & \multicolumn{3}{c}{IC13}                         & \multicolumn{3}{c}{IC15}                         & \multicolumn{3}{c}{MLT17}                         \\ \cmidrule(lr){2-4}\cmidrule(lr){5-7}\cmidrule(lr){8-10}
                               & R              & P              & F              & R              & P              & F              & R              & P              & F               \\ 
\midrule
ST-850k                        & \textbf{71.78} & 73.94          & 72.85          & 51.66          & 67.06          & 58.36          & \textbf{40.97} & 57.33          & 47.79           \\
UT-670k                        & 67.31          & \textbf{85.6}  & \textbf{75.36} & \textbf{53.06} & \textbf{81.69} & \textbf{64.33} & 39.07          & \textbf{66.73} & \textbf{49.28}  \\
LBTS-100k                      & 62.47          & 79.81          & 70.08          & 41.79          & 72.64          & 53.06          & 33.4           & 60.03          & 42.92           \\ 
\midrule
ST-850k + Real                 & 75.34          & 88.71          & 81.48          & 79.54          & 85.95          & 82.62          & 57.52          & 74.37          & 64.87           \\
UT-670k + Real                 & \textbf{76.8}  & 88.62          & 82.29          & \textbf{82.19} & 85.65          & 83.88          & \textbf{57.8}  & \textbf{74.32} & \textbf{65.02}  \\
LBTS-100k + Real               & 76.26          & \textbf{89.59} & \textbf{82.39} & 81.95          & \textbf{86.13} & \textbf{83.99} & 57.67          & 74.22          & 64.91           \\
\bottomrule
\end{tabular}
\end{table*}

\begin{table}[!t] 
\centering
\caption{Ablation Study: Qualitative Comparison between Different Configurations of Our Proposed Engine on ICDAR2015 Dataset Using EAST as the Baseline Detector.}
\label{tab5.4}
\begin{tabular}{cccc}
\toprule
\multirow{2}{*}{Train dataset} & \multicolumn{3}{c}{IC15}                                                          \\ \cmidrule(lr){2-4} 
                               & R                         & P                         & F                         \\ \midrule
w/o TLPNet                     & 35                        & 56.66                     & 43.27                     \\
w/o GTM                        & 31.68                     & 58.33                     & 41.06                     \\
w/o CHM                        & 37.41                     & 67.39                     & 48.11                     \\
w/o Postprocess                & 41.65                     & 67.21                    & 51.43                     \\
ALL                            & \textbf{42.95}            & \textbf{68.14}            & \textbf{52.69}            \\ \midrule
w/o TLPNet + Real              & 79.35                     & 86.92                     & 82.96                     \\
w/o GTM + Real                 & 80.36                     & \textbf{87.15}            & 83.62                     \\
w/o CHM + Real                 & 79.78                     & 85.85                     & 82.71                     \\
w/o Postprocess + Real         & \multicolumn{1}{r}{80.36} & \multicolumn{1}{r}{86.84} & \multicolumn{1}{r}{83.47} \\
ALL + Real                     & \textbf{81.03}            & 86.66                     & \textbf{83.75}            \\ \bottomrule
\end{tabular}
\end{table}

In addition, we created mixed synthetic datasets from different synthetic datasets to find out whether the data generated from different synthesis methods could play a complementary role during the training of the scene text detector. EAST was trained using the same configuration as the above experiment, and the evaluation results are summarized in Table~\ref{tab5.3-1}. Without using real data, EAST achieved the best F-score when trained on VISD-5k + UT-5k, which was higher than the results obtained with VISD-10k or UT-10k individually. However, this synergetic effect disappeared when it served as pretraining data. The performance of EAST trained on the VISD-5k + UT-5k + Real is almost in the range of that achieved with UT-10k + Real to VISD-10k + Real, which cannot surpass the better performance between UT-10k + Real and VISD-10k + Real. A similar approximately linear relationship can also be found in other mixed datasets, including LBTS. 
On the other hand. when the mixed data serve as the pretraining data, we found that EAST trained with VISD-5k + LBTS-5k + Real or UT-5k + LBTS-5k + Real, performed better than that trained with synthetic data from a single source, such as VISD-10k + Real or UT-10k + Real.

Finally, we generated 100k synthetic images to test the scalability of our LBTS. We compared LBTS-100k with the full-size ST \cite{gupta_synthetic_2016} and UT \cite{LongUnreal2020}. We trained EAST with 300,000 steps on different full-size datasets; the other configuration was the same as the above experiments. The evaluation results of EAST are presented in Table~\ref{tab5.3-2}. We observed that the performance of EAST improved when the number of generated datasets increased. Furthermore, EAST trained on LBTS-100k + Real achieved a competitive performance compared with that trained on ST-850k + Real and UT-670k + Real.


\subsection{Ablation Study} 
In this section, we investigated the effectiveness of different settings of the proposed data-generation engine. The text location proposal network (TLPNet), geometry transformation module (GTM), color harmonization module (CHM), and postprocessing were the focus. The evaluation results of the EAST trained on the datasets generated by different configurations on the ICDAR2015 dataset are reported in Table~\ref{tab5.4}.
\begin{itemize}

\item[\textbullet] \textbf{Text Location Proposal Network} Given a background image, TLPNet aims to propose suitable regions for text embedding, which are usually relatively plain areas, as depicted in Fig.~\ref{fig5.1.2}. To investigate its significance, we conducted an ablation study in which we replaced the output of TLPNet with an image, whose pixels value are all set to 1. This means the texts can appear at any location within the background image. The evaluation result, presented in Table~\ref{tab5.4} emphasizes that TLPNet improves the quality of the generated synthetic data whether they served as the sole training data or the pretraining data. 

\item[\textbullet] \textbf{Geometry Transformation Module} To assess the importance of the GTM, we replaced this module in our generation engine with a random transformation matrix generator. However, employing a completely random matrix generator is not advisable as it will heavily distort the text instances, resulting in extremely unrealistic results. For this reason, we adopted the random transformation matrix generator from a word-level SynthText engine \cite{Tangstroke2021} to reasonably transform the perspective of text instances, at least at the patch level. From Table~\ref{tab5.4}, firstly, we observed that data generated with GTM serves as better independent training data and pretraining data for the text detector. This reveals the importance of our GTM function in the synthesis engine. Secondly, we noticed that when using synthetic data solely for training, there is a substantial performance gap between datasets generated w/o GTM and ALL. Nevertheless, this gap significantly diminishes when we incorporate real data for fine-tuning. This phenomenon further supports the conclusion drawn in the last subsection, highlighting that lower performance in the pretraining model does not necessarily lead to low performance in the fine-tuned model.

\item[\textbullet] \textbf{Color Harmonization Module} To evaluate the advantages of the CHM, the color-deciding process in our engine was replaced with that of the SynthText engine \cite{gupta_synthetic_2016}, where the text color is determined by referencing a learned dictionary based on the background's local statistic information. We can observe that the performance of EAST decreased when our CHM was missing.

\item[\textbullet] \textbf{Post-processing} To confirm the contribution of post-processing of our engine, we generated a dataset without applying post-processing and evaluated the quality of this dataset. Table~\ref{tab5.4} implies that our post-processing techniques can enhance data diversity and improve the overall quality of generated data.
\end{itemize}

\subsection{Discussion} 
Based on our comprehensive experimental results, although we cannot explicitly determine the specific type of data that benefits the training of text detectors, we can summarize several findings that prior studies have not addressed.
First, we discovered that the performance of a text detector trained on both synthetic and real data is not strictly positively correlated with that trained only on synthetic data, even if the performance gap of the synthetic data is large. 
Second, the integration of different synthetic datasets generally improves the performance of the text detector; however, the extent of improvement differs based on the utilization of the mixed synthetic datasets. When using mixed synthetic data as independent training data, better performance can be achieved than that of datasets from a single source. However, when real data are involved in fine-tuning, the performance of the mixed data fails to surpass the best performance achieved by the single source dataset.


Our generation engine has several limitations. 
Firstly, the performance of TAANet, especially the GTM, is heavily influenced by the results of TLPNet. There exists a gap between the training data and inference data in TLPNet, where text-erased images usually have relatively large and flat areas with strong leading lines, such as the edges of signage or billboards, but the inference data are usually more diverse. A poor prediction of the text region often results in an unsatisfactory final output for human perception. This is because the  GTM struggles to reasonably transform the perspective of text instances when the leading lines are missing in the background image. 
Secondly, in our proposed TAANet, the forward process is based on one text instance, thus, our method neglects to model the relationship between text instances. We opted to abandon text instances that were too close or that intersected with other texts, as it is uncommon for text to overlap in the real world. However, this trick usually leads to a disorganized layout of text instances and a reduction in generation efficiency.
We believe that a unified training and generation structure may improve the generation results, and we expect future studies to successfully address these problems for learning-based scene-text image synthesis tasks.

\section{Conclusion} 
\label{Conclusion}
In this study, we first propose a new scene text dataset called DecompST, which can decompose real-world scene-text images into pure background images and pure text instances using text-erased images and stroke-level masks. Leveraging the DecompST dataset, we introduce a learning-based scene-text image synthesis engine, termed LBTS, which comprises a text location proposal network (TLPNet) and a text appearance adaptation network (TAANet). TLPNet is a segmentation network, capable of predicting suitable regions for text embedding. It is trained with the data pair of text-erased images and the mask of text regions, where text regions were extended from GT BBoxes based on appearance similarity and boundary information.
TAANet consists of a geometry transformation module and a color harmonization module. These components can adaptively adjust the perspective and color of the synthetic text instance to ensure compatibility with the background.
By combining our trained TLPNet and TAANet, we have developed a synthetic scene-text image generation engine and verified the effectiveness of our generated dataset using two popular baseline text detectors. Comprehensive experiments demonstrated the effectiveness of our proposed method in generating pretraining data for scene text detection.

\ifCLASSOPTIONcaptionsoff
  \newpage
\fi

\bibliographystyle{IEEEtran}
\bibliography{new_ref}

\begin{IEEEbiography}[{\includegraphics[width=1in,height=1.25in,clip,keepaspectratio]{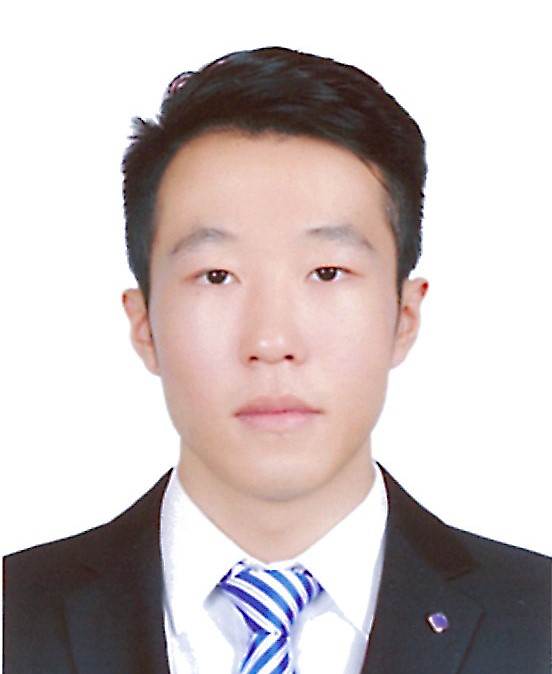}}]{Zhengmi Tang}
received his B.E. degree from Xidian University, Shaanxi, China, in 2017 and his M.E. degree in cybernetics engineering from Hiroshima University, Japan in 2020. He is currently pursuing a Ph.D. in communication engineering at the IIC-Lab at Tohoku University, Japan. His current research interests include computer vision, scene text detection, and data synthesis.
\end{IEEEbiography}

\begin{IEEEbiography}[{\includegraphics[width=1in,height=1.25in,clip,keepaspectratio]{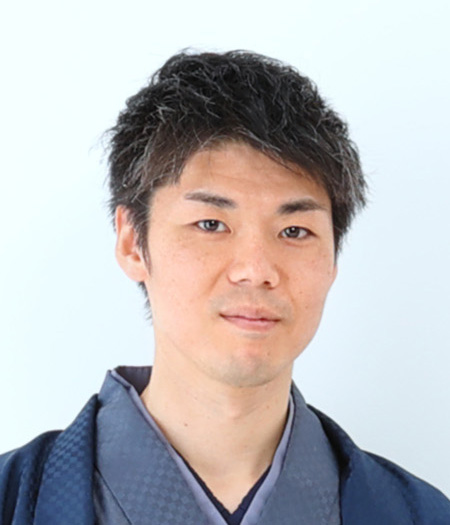}}]{Tomo Miyazaki}
 (Member, IEEE) received his B.E. and Ph.D. degrees from Yamagata University (2006) and Tohoku University (2011), respectively. From 2011 to 2012, he worked on the geographic information system at Hitachi, Ltd. From 2013 to 2014, he worked at Tohoku University as a postdoctoral researcher. Since 2015, he has been an Assistant Professor at the university. His research interests include pattern recognition and image processing.
\end{IEEEbiography}

\begin{IEEEbiography}[{\includegraphics[width=1in,height=1.25in,clip,keepaspectratio]{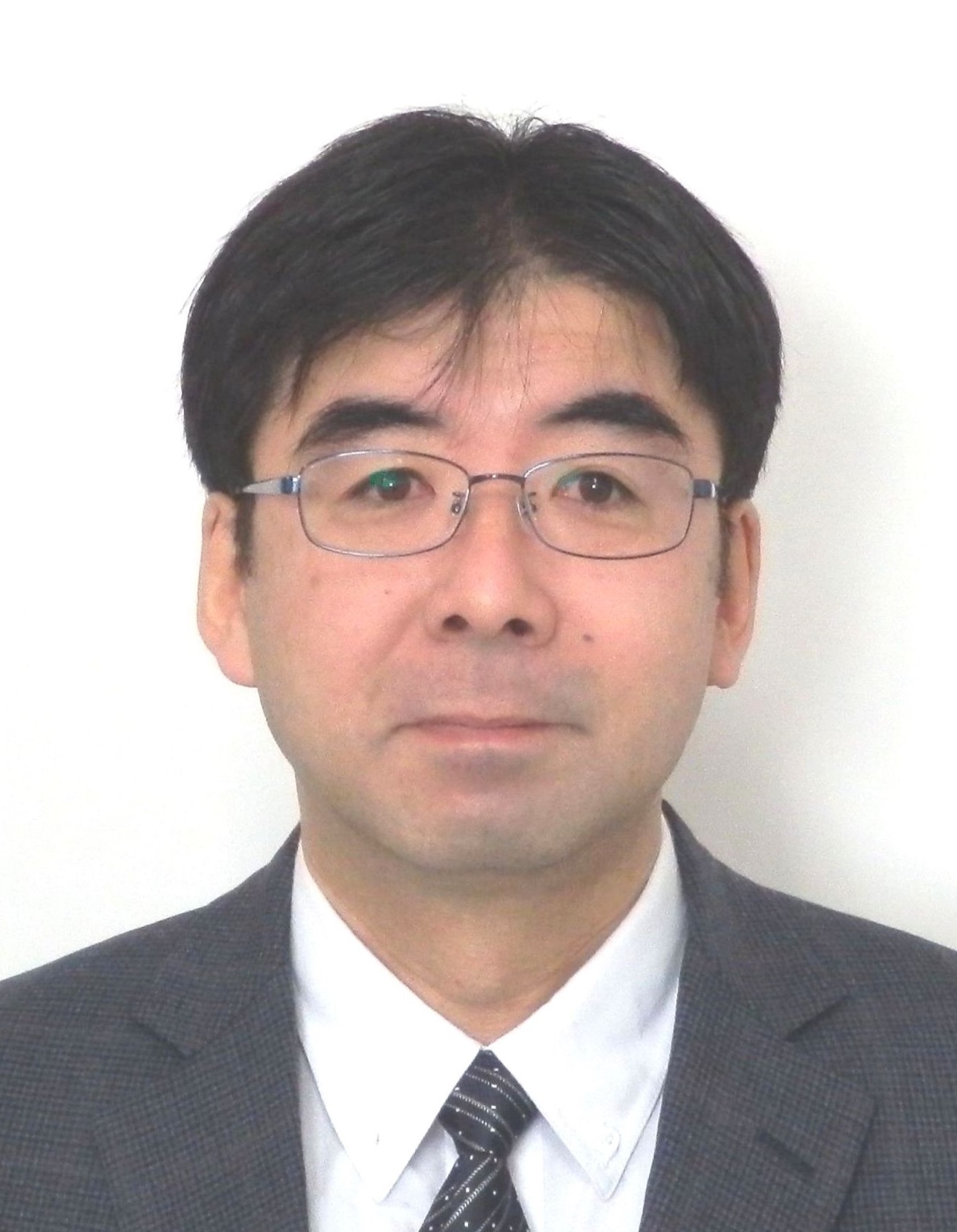}}]{Shinichiro Omachi}
(M’96-SM’11) received his B.E., M.E., and Ph.D. degrees in Information Engineering from Tohoku University, Japan, in 1988, 1990, and 1993, respectively. He worked as an Assistant Professor at the Education Center for Information Processing at Tohoku University from 1993 to 1996. Since 1996, he has been affiliated with the Graduate School of Engineering at Tohoku University, where he is currently a Professor. From 2000 to 2001, he was a visiting Associate Professor at Brown University. His research interests include pattern recognition, computer vision, image processing, image coding, and parallel processing. He served as the Editor-in-Chief of IEICE Transactions on Information and Systems from 2013 to 2015. Dr. Omachi is a member of the Institute of Electronics, Information and Communication Engineers, the Information Processing Society of Japan, among others. He received the IAPR/ICDAR Best Paper Award in 2007, the Best Paper Method Award of the 33rd Annual Conference of the GfKl in 2010, the ICFHR Best Paper Award in 2010, and the IEICE Best Paper Award in 2012. He served as the Vice Chair of the IEEE Sendai Section from 2020 to 2021.
\end{IEEEbiography}

\end{document}